# Out-of-distribution Reject Option Method for Dataset Shift Problem in Early Disease Onset Prediction

Taisei Tosaki[1], Eiichiro Uchino[1], Ryosuke Kojima[1], Yohei Mineharu[1], Yuji Okamoto[1], Mikio Arita[2], Nobuyuki Miyai[2], Yoshinori Tamada[3], Tatsuya Mikami[4], Koichi Murashita[5], Shigeyuki Nakaji[4], Yasushi Okuno[1,*]

[1] Graduate School of Medicine, Kyoto University, Japan

[2] Graduate School of Health and Nursing Science, Wakayama Medical University, Japan

[3] Research Center for Health-Medical Data Science, Graduate School of Medicine, Hirosaki University, Japan

[4] Innovation Center for Health Promotion, Graduate School of Medicine, Hirosaki University, Japan

[5] Center of Innovation Research Initiatives Organization, Hirosaki University, Japan

[*] okuno.yasushi.4c@kyoto-u.ac.jp

## Abstract：

Machine learning is increasingly used to predict lifestyle-related disease onset using health and medical data. However, its predictive accuracy for use is often hindered by dataset shift, which refers to discrepancies in data distribution between the training and testing datasets. This issue leads to the misclassification of out-of-distribution (OOD) data. To diminish dataset shift in real-world settings, this paper proposes the out-of-distribution reject option for prediction (ODROP). This method integrates an OOD detection model to preclude OOD data from the prediction phase. We used two real-world health checkup datasets (Hirosaki and Wakayama) with dataset shift, across three disease onset prediction tasks: diabetes, dyslipidemia, and hypertension. Both components of ODROP method—the OOD detection model and the prediction model—were trained on the Hirosaki dataset. We assessed the effectiveness of ODROP on the Wakayama dataset using AUROC-rejection rate curve plot. In the five OOD detection approaches (the variational autoencoder, neural network ensemble std, neural network ensemble epistemic, neural network energy, and neural network gaussian mixture based energy measurement), the variational autoencoder method demonstrated notably higher stability and a greater improvement in AUROC. For example, in the Wakayama dataset, the AUROC for diabetes onset increased from 0.80 without ODROP to 0.90 at a 31.1% rejection rate, and for dyslipidemia, it improved from 0.70 without ODROP to 0.76 at a 34% rejection rate. In addition, we were able to categorize dataset shifts into two types using SHAP clustering—those that considerably affect predictions and those that do not. We expect that this classification will help standardize measuring instruments. This study is the first to apply OOD detection to actual health and medical data, demonstrating its potential to substantially improve the accuracy and reliability of disease prediction models amidst dataset shift.

## Introduction：

Advancements in machine learning have made it possible to predict disease risk based on large-scale multivariate health and medical data[1–4]. Machine learning models for disease onset prediction, especially those based on lifestyle, diet, and exercise habits, are expected to individually prevent diseases by forecasting the potential development of lifestyle-related diseases, such as diabetes and hypertension, by presenting individual contributing factors[5]. Constructing higher-performing machine learning models requires a vast amount of training data. Hence, multi-item health and medical data are accumulated worldwide from patients with chronic diseases and healthy individuals alike[6–8].

The difficulty of data sharing and scarcity of health and medical data emphasize the importance of using a disease onset prediction model built on health checkup data collected at one site for use at other sites. In this case, the disease onset prediction model faces the challenge of dataset shift[9–11], a problem where the probability distributions of training and test data differ (Fig. 1(a)), causing the test data to have in-distribution (ID) and out-of-distribution (OOD) data. The distribution difference means that one of the model assumptions, that is, training and test data distributions are equal, does not hold, leading to the model's misclassification of the OOD test data. The problem arises when the data acquisition location for training and actual testing differ[9,11]. Factors affecting the dataset shift problem include regional differences in diet, lifestyle, and exercise habits, as well as discrepancies in the measurement instruments used at various sites. Such variations based on unique regional characteristics make it challenging to avoid dataset shift.

In prior research, most early disease onset prediction models have not addressed the dataset shift. These models have been evaluated solely on the internal dataset[1–4] or reported a reduction of area under the receiver operating curve (AUROC) when applied to different sites due to OOD data[12]. Some studies[13,14] have attempted to enhance robustness of disease onset predictions for external datasets by excluding OOD data from prediction samples. They used conformal prediction[15], a statistical testing method using features transformed by a neural network encoder to identify OOD data, and then made predictions on ID data. As a result, the OOD detection and prediction models become integrated, limiting the ability to use pre-trained predictive models with high discriminative power. On the other hand, OOD detection models that operate independently of prediction models are quite popular in computer vision[16–20]. However, unlike images, where human experts can annotate ID and OOD data, electronic health records and health checkup data present challenges for such annotation. This difference makes it difficult to directly apply OOD detection models in computer vision to these types of data. Thus, methods for effectively handling OOD health and medical data derived from dataset shift remain insufficient.

The aim of this study is to explore effective methods to address the dataset shift problem in disease onset prediction models when testing health and medical data with different distributions from the training data. Our proposed approach involves a two-stage predictive method called out-of-distribution reject option for prediction (ODROP, Fig. 1(b)), which uses an OOD detection model to reject OOD data from a test dataset. In the first stage, an OOD detection model score the divergence between the training and test data distributions to discern the appropriateness of the test data as ID or OOD data. In the second stage, we include an option to avoid predictions for data identified as OOD. Our ODROP method derives from the known reject option method, which avoids class prediction when the classification confidence is within a certain range[21,22].

We refine this reject option method for OOD data caused by a dataset shift.

We used five OOD detection methods and two health checkup datasets with a dataset shift and evaluated their methods' effectiveness in three disease onset prediction tasks, namely diabetes, hypertension, and dyslipidemia, within one year. Our evaluation considered three aspects: stability, extent of improvement in the prediction performance metrics, and the proportion of rejected samples at maximum improvement. We identified the ODROP method using a variational autoencoder (VAE) [23] as the optimal OOD detection model. In addition, we compared the patterns of prediction contribution (SHAP) [24] values between the ID and rejected OOD data groups. We discovered for the first time that the dataset shift could be classified into those considerably contributing to disease onset prediction and those that do not.

Our contributions are following: This study is the first to apply out-of-distribution (OOD) detection models to real-world health and medical data, demonstrating their effectiveness in detail. Additionally, our proposed ODROP method offers a solution to the dataset shift problem. It enhances the robustness of existing clinical prediction models against dataset shift without altering their prediction mechanisms, thereby providing a practical and impactful advancement in the field.

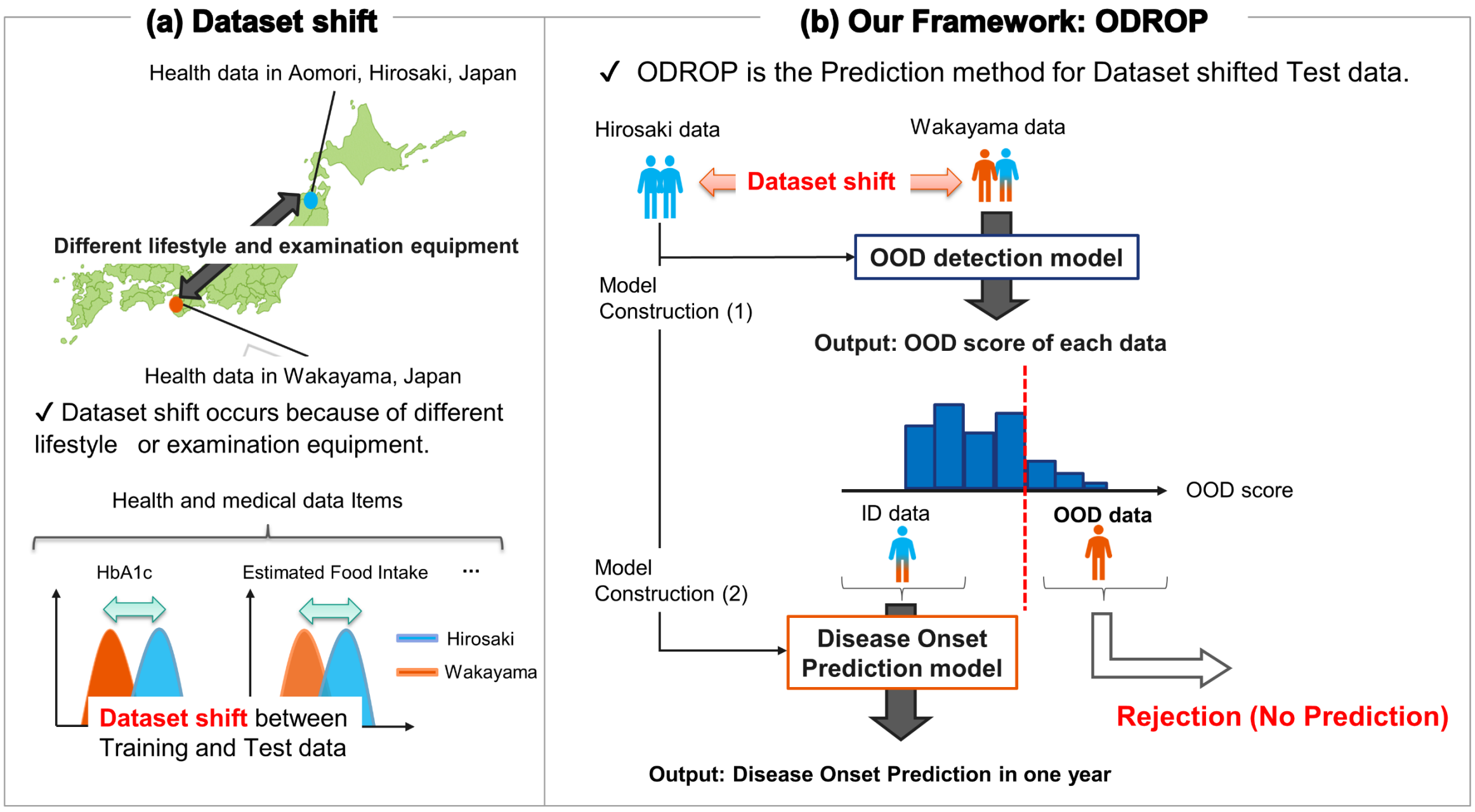

**Fig. 1 Overview of this study.**

**(a) Dataset shift**

This study used health checkup data from Hirosaki City in Aomori Prefecture, Japan, and Wakayama Prefecture, Japan, with dataset shift. The disease onset prediction model constructed from Hirosaki data has a lower prediction performance in Wakayama data than that of Hirosaki data due to the dataset shift.

**(b) Proposed Method—Out-of-distribution reject option for prediction; ODROP**

In the proposed method, an out-of-distribution (OOD) detection model constructed from Hirosaki health checkup data first calculates the OOD score of each Wakayama health checkup data. The OOD score represents suitability as OOD data. Thus, data with an OOD score above a threshold are classified as OOD data (right side of OOD score histogram). Finally, a disease onset prediction model constructed from Hirosaki data predicts the in-distribution (ID) data, which are appropriate for prediction.

## Results:

### Dataset shift between two Health checkup datasets

Several cohort studies[8,25,26] have been conducted that reflect the regional characteristics of Japan. Some of these studies have multi-item health examination data, including physiological and biochemical data, such as blood and respiratory metrics; data on personal activities, such as diet, exercise habits, and daily stress; and socioeconomic data, such as educational background and work environment. In this study, we used two multi-item health checkup datasets from different regions of Japan: Hirosaki City in Aomori Prefecture[8] and Wakayama Prefecture[25,26]. We conducted statistical tests to confirm dataset shifts between the two and plotted kernel density estimation (KDE) for each item. The results are presented in Table 1 and Fig. 2. Complete summary statistics for all items from both sites and the results of the statistical tests between the two sites can be found in Supplementary Table 1. The KDE plots in Fig. 2 visualize the distribution shifts in two health datasets. However, the overlapping regions in the distributions suggest that the Wakayama health checkup data (Wakayama data) can be divided into two groups, with one group having similar characteristics to the Hirosaki health checkup data (Hirosaki data).

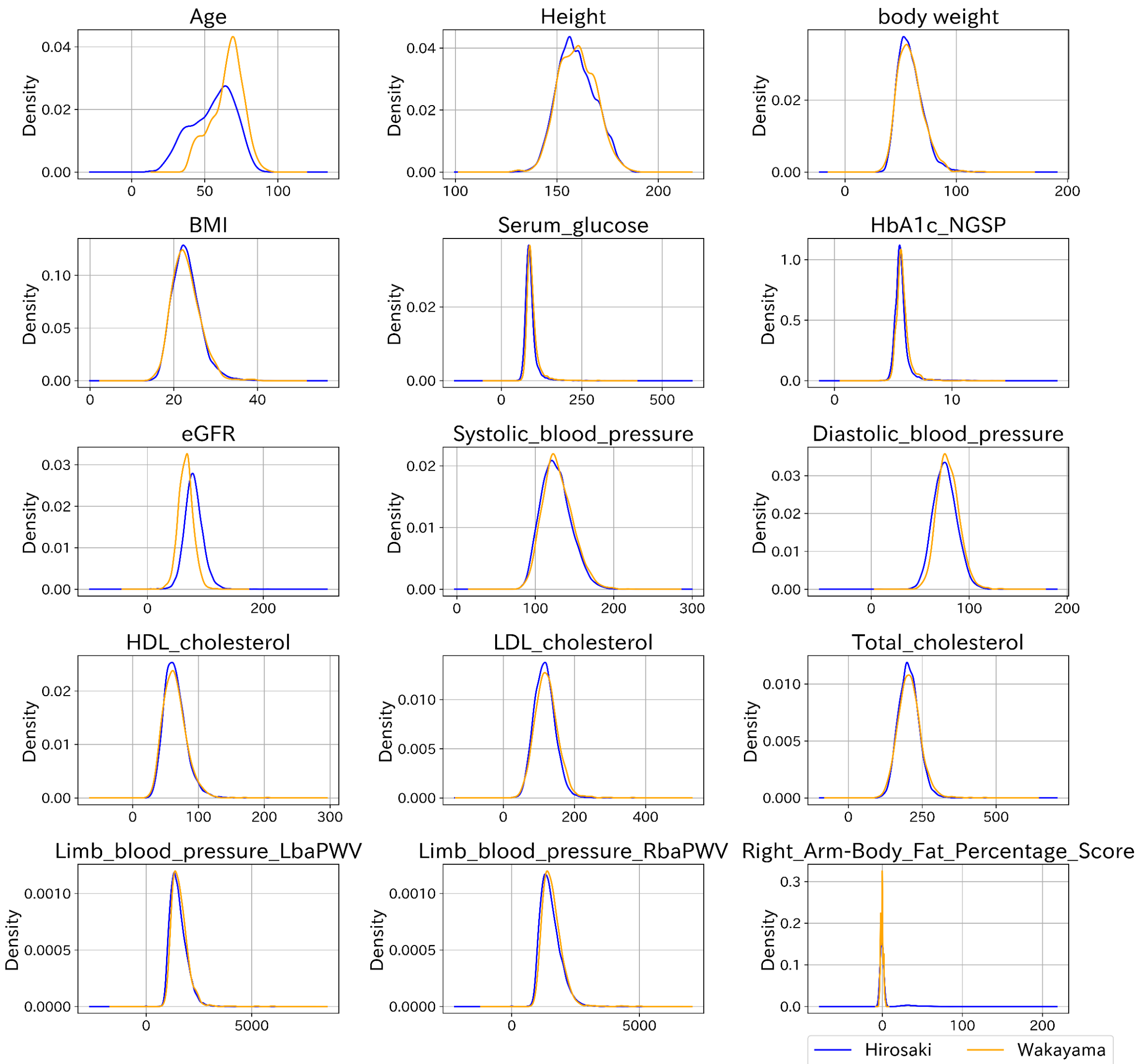

**Fig. 2 Kernel density estimation plot in the main items of Hirosaki and Wakayama health checkups**

**Baseline Evaluation of Hirosaki and Wakayama Health Checkup Data**

We confirmed the occurrence of the dataset shift problem: whether the predictive performance metrics in the Wakayama health checkup data decreased compared to the Hirosaki health checkup data, which is the training base for the disease onset prediction models. We compared the mean receiver operating characteristic (ROC) curve from 5-fold cross-validation at Hirosaki with the ROC curve for the Wakayama health checkup data in Fig. 3. The precision-recall (PR) curves are shown in Supplementary Fig. 1. The Wakayama health checkup AUROC is lower in the three disease onset prediction tasks compared to the Hirosaki mean AUROC, with decreases of 0.11 for diabetes, 0.09 for dyslipidemia, and 0.02 for hypertension. Similarly, PRAUC decreased for all tasks by 0.116, 0.253, and 0.012 for diabetes, dyslipidemia, and hypertension, respectively. Hypertension has the smallest decline in AUROC and PRAUC values. Hereafter, the mean AUROC from 5-fold cross-validation is referred to as the Hirosaki AUROC baseline, and that from the Wakayama health checkup data is referred to as the Wakayama AUROC baseline (the same applies to PRAUC).

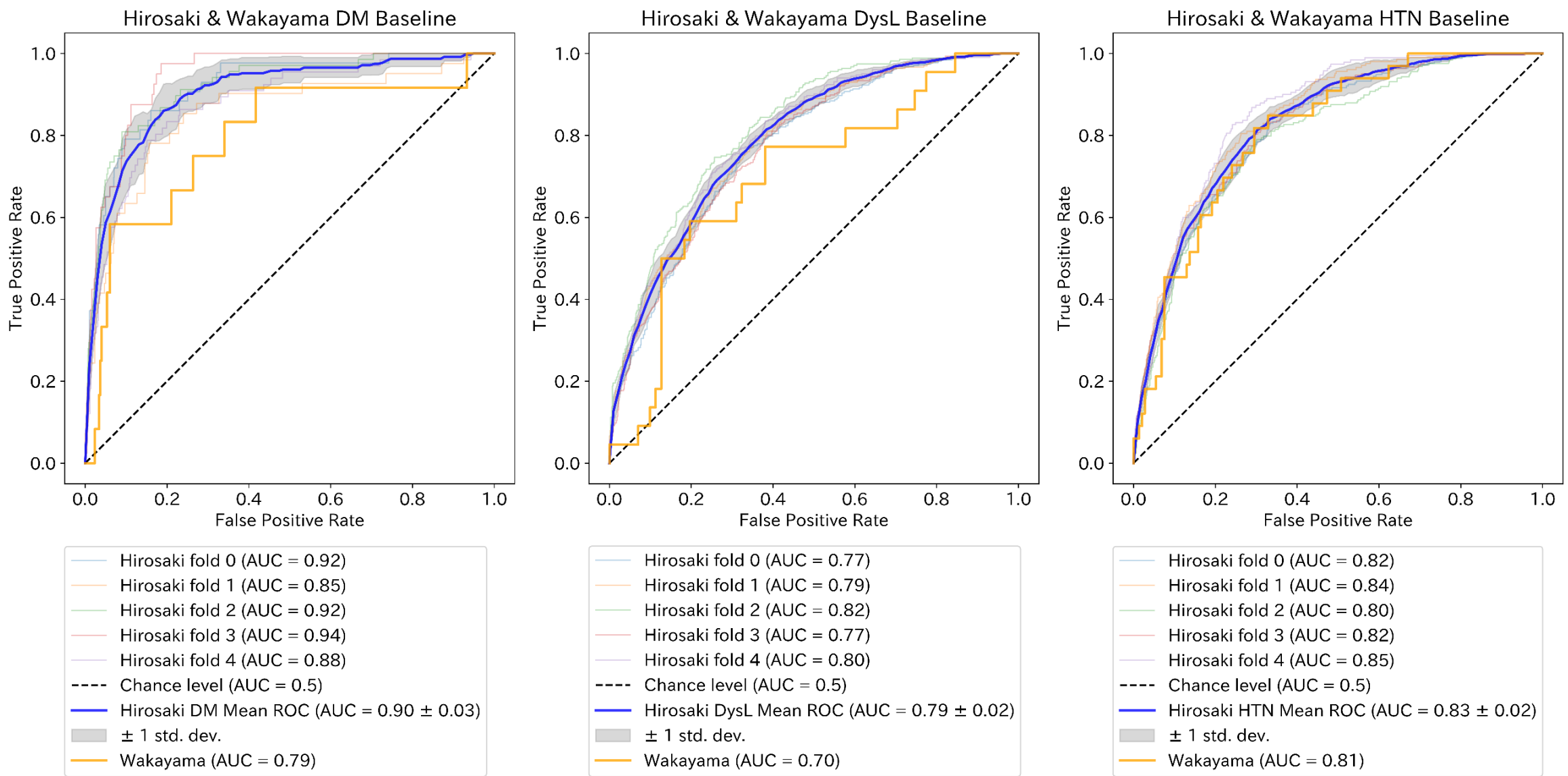

**Fig. 3 Comparison of AUROC baselines between Hirosaki and Wakayama health checkups**

The results of the 5-fold cross-validation ROC curves for each disease onset prediction task conducted in Hirosaki, along with their mean ± std ROC curves, compared to the ROC curve results from Wakayama health checkup data. The values in parentheses represent the AUROC values.

(Left): Prediction of diabetes onset within one year

(Center): Prediction of dyslipidemia onset within one year

(Right): Prediction of hypertension onset within one year

**Rejection Rate Evaluation**

We used the rejection rate for ODROP evaluation in health and medical data, which is the proportion of OOD data rejected from all test data. We assessed five OOD detection methods: VAE reconstruction loss (VAE reconstruction)[27], neural network ensemble std (ensemble std)[28], neural network ensemble epistemic (ensemble epistemic)[28], neural network energy (energy)[29], gaussian mixture based energy measurement (GEM)[30] for diabetes, hypertension, and dyslipidemia onset prediction within one year. The rejection curve[31] evaluates the extent of prediction metric improvement (AUROC or PRAUC on the y-axis) with the rejection rate (x-axis). The 0% rejection rate represents "baseline," which is the prediction metric value for all the test data. The result of 0% rejection rate, indicating the absence of ODROP method, corresponds to prediction metrics reported in previous studies. Increasing the rejection rate from 0% allows for the gradual exclusion of the OOD test data. We confirmed that subsequent exclusion led to a stepwise improvement in the predictive performance metrics of the model. In addition, to evaluate the stability of the prediction metric improvement when increasing the rejection rate, we evaluated the rank correlation coefficient between the prediction performance metric and rejection rate. The rank correlation coefficient is positive if the ODROP method improves the prediction performance metrics from the baseline at an increased rejection rate. In addition, the larger the coefficient, the more stable and consistent the improvement.

**Internal Validation using Hirosaki Health Checkups**

For internal validation, we used the proposed ODROP method on Hirosaki health checkup data, which do not exhibit a dataset shift, and evaluated it using 5-fold cross validation. The results for the AUROC across the three disease onset prediction tasks are shown in Fig. 4, and the PRAUC results in Supplementary Fig. 2.

From the bar graphs showing the rank correlation coefficients in Fig. 4a, we confirmed that VAE reconstruction was positive for diabetes; energy and ensemble std were positive for dyslipidemia; and GEM, energy, ensemble std, and VAE reconstruction were positive for hypertension. In Fig. 4b, the methods that improved the mean AUROC from the baseline were VAE reconstruction for diabetes; ensemble epistemic, ensemble std, and VAE reconstruction for dyslipidemia; and GEM, ensemble epistemic, ensemble std, and VAE reconstruction for hypertension. This indicates that these methods effectively improve the prediction performance metrics when rejecting OOD data. The method that showed the greatest improvement in mean AUROC from baseline was VAE reconstruction for diabetes and dyslipidemia and ensemble epistemic for hypertension. The maximum mean AUROC is 0.916 (rejection rate: 24.0%), 0.808 (33.2 %), and 0.848 (38.4 %) for diabetes, dyslipidemia, and hypertension, respectively. The maximum extent of AUROC improvement was 0.015 for diabetes, 0.017 for dyslipidemia, and 0.021 for hypertension. VAE reconstruction was the only method that indicated a tendency for AUROC improvement across the three disease onset prediction tasks.

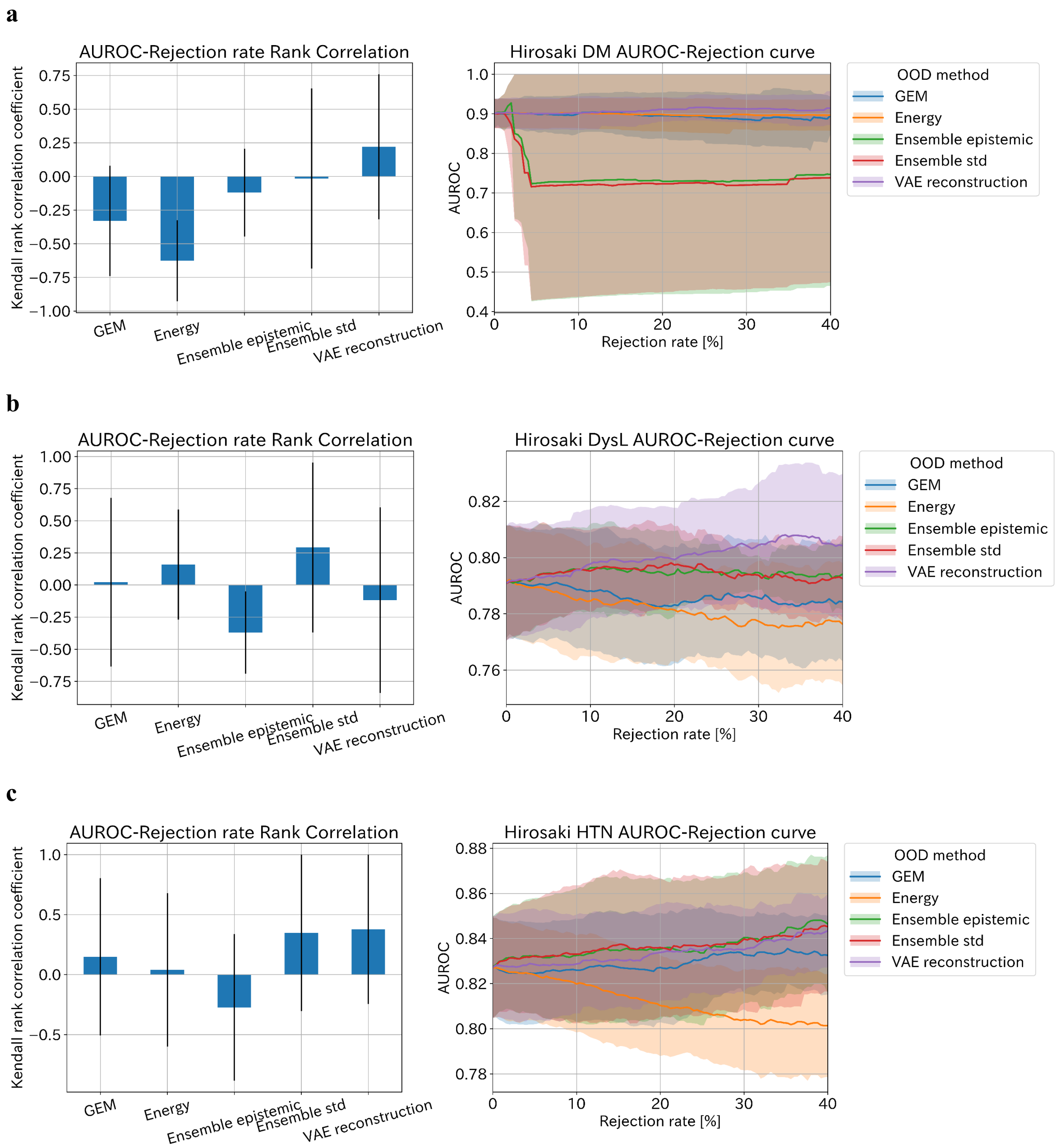

**Fig. 4 AUROC-rejection rate rank correlation coefficients and AUROC-rejection curves in Hirosaki health checkup**

**a: Diabetes Melius (DM), b: Dyslipidemia (DysL), c: Hypertension (HTN).**

Left Bar Plot: The mean ± std of rank correlation coefficient between rejection rate and AUROC.

Right Plot: AUROC-rejection curve. Y-axis is AUROC value (mean ± std) and x-axis is rejection rate.

In **a** and **c**, VAE reconstruction method showed a positive and considerable rank correlation coefficient, indicating a nearly monotonic improvement trend. VAE reconstruction method also demonstrated the greatest improvement from the baseline AUROC at a 0% rejection rate in **a** and **b**. **c** showed an improvement extent nearly equivalent to that of ensemble epistemic method, which had the largest improvement range.

**External Validation using Wakayama Health Checkups**

We used five OOD detection methods, namely VAE reconstruction, ensemble epistemic, ensemble std, energy, and GEM, and applied each ODROP approach to the Wakayama health checkups, which had a dataset shift between the Hirosaki health checkups. The results for the AUROC across the three disease onset prediction tasks are shown in Fig. 5, and the PRAUC results in Supplementary Fig. 3. For diabetes and dyslipidemia, VAE reconstruction method yielded positive rank correlation coefficients for the AUROC. The ensemble epistemic and ensemble std method were positive for hypertension. VAE reconstruction method also demonstrated positive rank correlations for PRAUC in diabetes and hypertension, suggesting it consistently improved the predictive performance metrics.

In Fig. 5, only VAE reconstruction method is shown to improve AUROC for diabetes, reaching a peak of 0.90 at 31.1% rejection rate, marking a 0.1 improvement over the Wakayama baseline. For dyslipidemia, VAE reconstruction method improved AUROC at a lower rejection rate than the ensemble epistemic, maintaining around 0.75 and peaking at 0.76. For hypertension, methods using neural network ensembles, ensemble std and epistemic show similar improvements in AUROC, with VAE reconstruction method maintaining near-baseline performance. In the three diseases investigated, the energy method, which was initially developed for image-based OOD detection, did not improve the AUROC scores but progressively improved the PRAUC scores and is a notable finding of this study. Additionally, the GEM method, an advanced version of the energy model, consistently underperforms the energy method in both predictive performance metrics. This indicates that the advancements in image-domain methods do not always correlate with improved outcomes.

These findings suggest that VAE reconstruction is the most suitable OOD detection method for the ODROP approach because of its considerable improvement in predictive performance metrics, lower rejection rates during improvement, and stable enhancement across various rejection rates, particularly during gradual increases in the rejection rate.

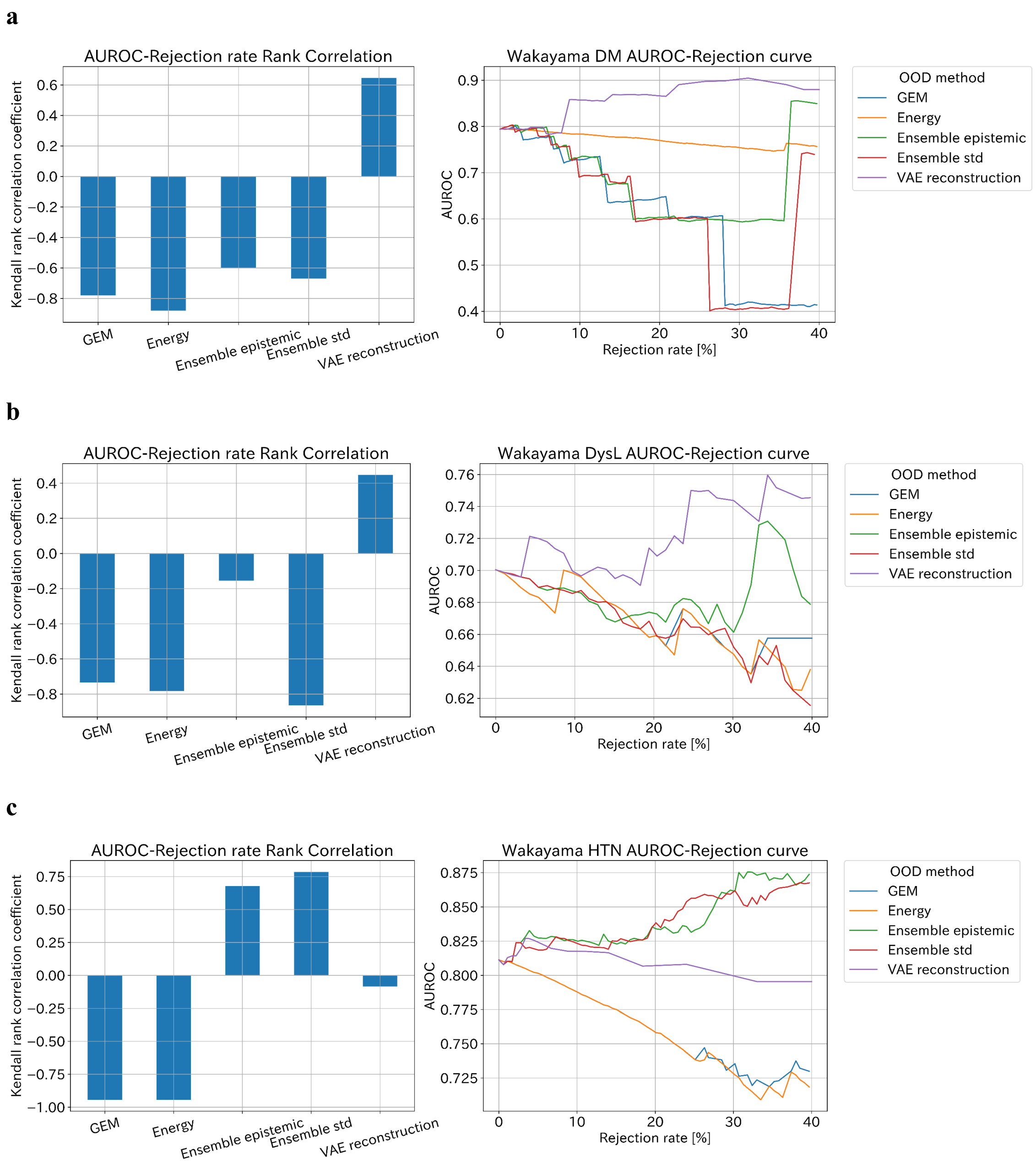

**Fig. 5 AUROC-rejection rate rank correlation coefficients and AUROC-rejection curves in Hirosaki health checkup**

**a: Diabetes Melius (DM), b: Dyslipidemia (DysL), c: Hypertension (HTN).**

VAE reconstruction was the only method with a positive rank correlation coefficient in **a** and **b**, showing a stable improvement in AUROC through the rejection curve. In **c**, ensemble epistemic and ensemble std had positive coefficients, with the rejection curve confirming an upward trend in AUROC.

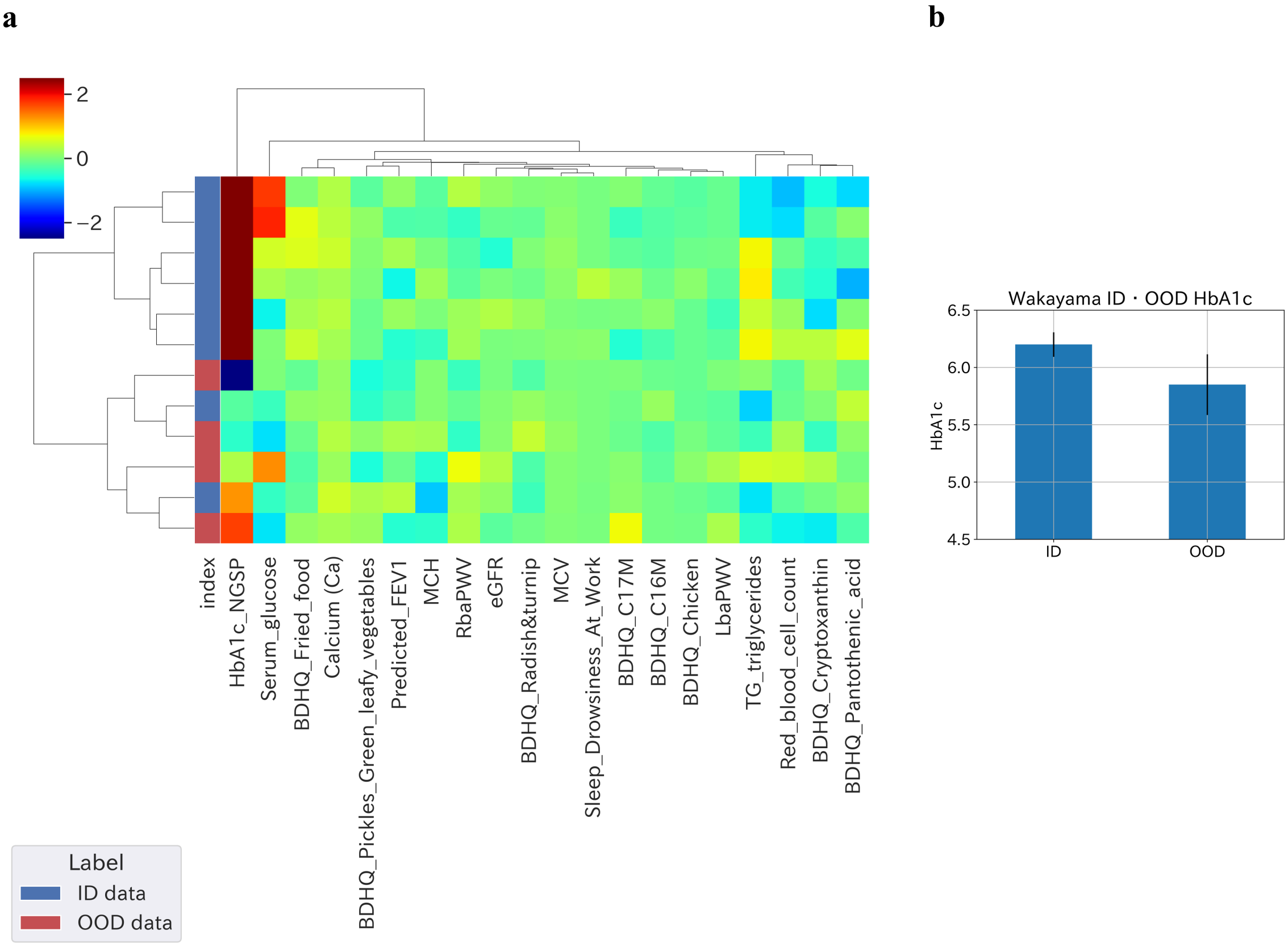

**Fig. 6 Dataset shift in diabetes onset within one-year records for diabetes onset prediction model**

**a. SHAP clustering for diabetes onset within one-year records in Wakayama health checkups**

This figure shows a hierarchical clustering analysis using SHAP values from a one-year diabetes onset prediction model for individuals from Wakayama health checkup data who developed diabetes within one year. A colormap represents the magnitude of the SHAP values calculated by the prediction model, with the vertical axis listing the Wakayama health checkup data of individuals who developed diabetes within one year. The horizontal axis without an index column shows the names of each examination item used in the prediction model, whereas an index column is IDs and OOD labels based on the VAE reconstruction loss threshold at the rejection rate of 31.1%, where AUROC was maximized in the rejection curve.

**b. HbA1c Levels in one-year diabetes onset Wakayama ID and OOD data (mean±std)**

The HbA1c value, which showed the most pronounced pattern differences between ID and OOD in SHAP Clustering, was presented as mean ± std for both ID and OOD data.

To identify the items that considerably impact disease onset prediction owing to the dataset shift, we used SHAP[24] values, which quantitatively represent the contribution of each predictor to the model's output. Differences in the SHAP value patterns between the ID and OOD data groups, can help determine which items cause considerable dataset shifts that affect disease onset prediction.

We show the clustering result using VAE reconstruction as an OOD detection method for ODROP method in predicting diabetes onset within one year (Fig. 6 a). The clustering of each item was split into two clusters: one with a high tendency for absolute SHAP values, notably HbA1c, and the other with lower values across the remaining items. The clustering of each record for diabetes onset within one year was split into two groups based on the HbA1c SHAP values, which were identified as the ID and OOD data groups based on the labels assigned. The actual HbA1c values for the ID and OOD groups (Fig. 6b), reveal that the OOD group has relatively lower HbA1c levels than the ID group. Thus, this dataset shift in HbA1c is considerable for the model predicting diabetes onset within one year. The results of SHAP clustering for individuals diagnosed with dyslipidemia or hypertension within a year of the Wakayama health checkup data are provided in Supplementary Figures 4A and B, respectively.

## Discussion:

This study demonstrates that the proposed ODROP method can improve predictive performance metrics from the baseline in disease onset predictions across two health checkup datasets with different regional characteristics within the same country. This approach offers a viable solution to the dataset shift problem by addressing the issue of discrepancies between the predictive performance at the model training location and the actual application site[9,11]. Evaluation of the three perspectives revealed that the ODROP method using VAE reconstruction as the OOD detection method was optimal. In addition, we analyzed the SHAP value patterns of the disease onset prediction model and discovered, for the first time, that datasets from different regions included dataset shifts that considerably impacted disease onset prediction and those that did not.

We showed that the ODROP method could improve the prediction metrics of diabetes, dyslipidemia, and hypertension onset within one year when using the Wakayama and Hirosaki health checkup data as the test and training data, respectively. The VAE reconstruction for diabetes prediction and ensemble epistemic ODROP method for hypertension prediction considerably improved the AUROC scores, reaching 0.90 and 0.875, respectively. These improvements matched or exceeded Hirosaki's baseline performance. Thus, the ODROP method can adequately address the dataset shift problem in disease onset prediction within one year. These results also suggest that the Wakayama health checkup data, affected by dataset shifts, contained groups similar and dissimilar to the Hirosaki health checkup data. The ODROP method effectively isolates and predicts similar groups, improving the predictive metric performance. This indicates the potential effectiveness of the ODROP method in other regions with test datasets comprising groups similar and dissimilar to the training data, providing a viable solution to the dataset shift problem in health data analytics.

Internal and external validations were conducted to explore the most appropriate OOD detection method for health and medical data using the ODROP method. Internal validation demonstrated improved predictive performance metrics for all three disease onset predictions. The VAE reconstruction ODROP method showed superior stability and magnitude of improvement in the AUROC, suggesting its effectiveness even when applied within the same location as the training dataset. In the external validation, the VAE reconstruction ODROP method uniquely and consistently improved the AUROC for diabetes and dyslipidemia onset predictions, although it maintained the AUROC baseline for hypertension onset prediction within one year. These results suggest VAE reconstruction as the most effective and optimal OOD detection method in the ODROP approach for health and medical data, considering its stable improvement in predictive performance metrics and considerable improvement range. As an unsupervised learning model that does not require a target variable, VAE allows for flexible applications across multiple prediction tasks without retraining the neural network classifier for each task. This versatility gives the VAE an advantage over neural network classifier-based OOD detection methods (ensemble epistemic, ensemble std, energy, and GEM), enabling more efficient deployment of the ODROP approach across various predictive scenarios. Energy and GEM, initially developed for image-based OOD detection, underperform compared with other methods in structured data, including health and medical data. The lack of superior results suggests that image-based OOD detection models do not always translate well to structured data. This highlights the need for new benchmarks tailored to structured datasets, particularly health and medical datasets.

The proposed method has two advantages. First, the OOD detection model operates independently of the predictive model. This allows for the straightforward addition of an OOD detection model to existing medical or clinical prediction models using structured data, facilitating improvements without modifying existing prediction models. This integration can also address dataset shift and provide more reliable prediction outcomes without altering the original models. Second, the ODROP method does not require dataset sharing between training and testing sites when constructing the OOD detection model. Previous approaches to addressing dataset shift assumed simultaneous access to training and test data[32,33], a challenging requirement for health and medical data owing to privacy concerns. Thus, the ODROP method is a practical solution to address dataset shift without data sharing.

Furthermore, we compared the SHAP clustering patterns of item contributions between the ID and OOD groups in patients who developed diabetes within one year. Dataset shifts can be classified into two: those that considerably impact predictions and those that do not. Previous studies have systematized dataset shifts by starting with a covariate shift[10]. In contrast, this study is the first to focus on dataset shifts in terms of their contribution to the prediction model. Identifying items that cause considerable dataset shifts for predictive models is crucial because these identifications could lead to the standardization of measurement instruments across multiple hospital sites and practical measures for addressing dataset shifts.

One limitation of the proposed ODROP method is that it cannot provide prediction results for all test data and requires predictive models optimized for data from each testing site. Although domain adaptation and generalization techniques[34,35] have been explored for constructing such models, they require retraining neural network models, necessitating large sample sizes and data sharing across sites for fine-tuning. Thus, the selection or combination of these techniques or our method for appropriate manner is of importance to achieve effective prediction in clinical settings.

The development of the ODROP method employing an OOD detection model enabled reliable and accurate predictions across health and medical datasets affected by dataset shift. This study first evaluated multiple

OOD detection methods in health and medical data, assessing improvements in predictive performance metrics considering stability, magnitude, and rejection rate in three disease onset prediction tasks. Accordingly, we demonstrated that VAE reconstruction is the optimal OOD detection method for health and medical data. Our ODROP method provides a general solution to the dataset shift problem because it enhances the robustness of existing clinical prediction models against dataset shift without modifying the prediction mechanism.

## Methods:

### Data

We used health checkup data from the Iwaki Health Promotion Study[8] from 2005 to 2020 and the Wakayama Study[25,26] from 2018 to 2019. These datasets are comprehensive, encompassing over 2000 items, including physiological and biochemical data such as blood and respiratory metrics, personal lifestyle data such as diet and stress, and socioenvironmental data such as education and employment, showcasing diverse regional characteristics within Japan. Of the 383 common items between the two datasets, we selected 334 items with less than 50% missing data in both datasets and had data available for at least two consecutive years (Fig. 7). We conducted statistical tests between the Hirosaki and Wakayama health checkup data across 334 items and 3 additional items representing labels indicating the onset of diabetes, dyslipidemia, and hypertension within one year. For continuous variables, we used Welch's t-test, whereas for discrete variables, we used the $\chi 2$ test and Fisher's exact test following Cochran's rule. This study was approved by the Hirosaki University Faculty of Medicine Ethics Committee (annual approval, latest approval number: 2023-007-1) and conducted in accordance with the Declaration of Helsinki. Written informed consent was obtained from all participants.

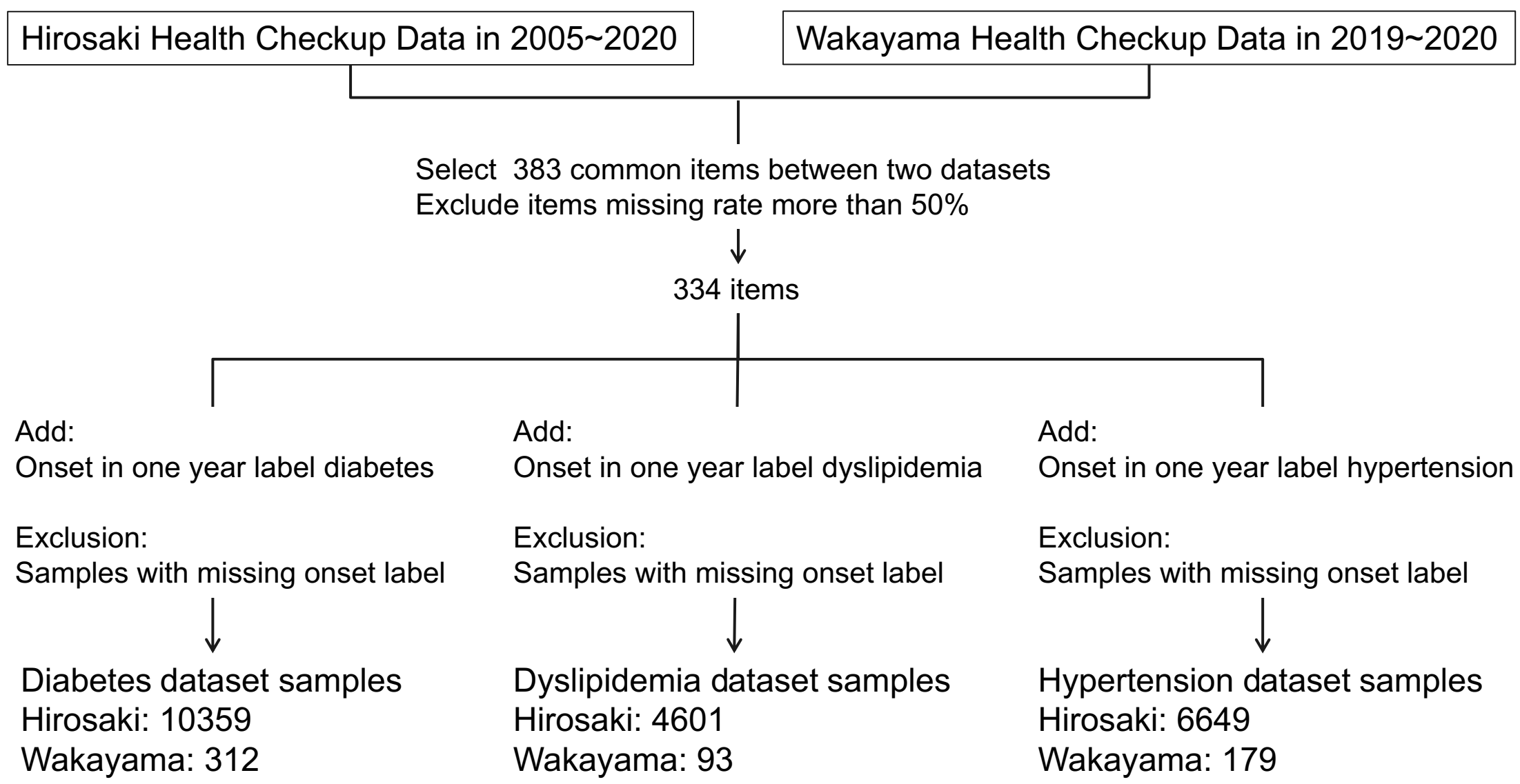

**Fig. 7 Data Preprocessing Flow**

### ODORP Flow

**Algorithm 1** ODROP Flow

**Require:** Training dataset $\mathcal{D}_{\text{train}} = \{(\boldsymbol{x}_i, y_i)\}_{i=1}^{N_{\text{train}}}$, Test dataset $\mathcal{D}_{\text{test}} = \{(\boldsymbol{x}_j, y_j)\}_{j=1}^{N_{\text{test}}}$, OOD score threshold $\tau_{\text{threshold}}$

**Ensure:** Prediction results for in-distribution test samples; Out-of-distribution samples are rejected.

**// Training Phase**
**if** Using classifier-based OOD detection **then**
    OOD_Model.fit($\mathcal{D}_{\text{train}}$)
**else**
    OOD_Model.fit($\{\boldsymbol{x}_i\}_{i=1}^{N_{\text{train}}}$) ▷ Density-based OOD detection
**end if**
Prediction_Model.fit($\mathcal{D}_{\text{train}}$)
**// Inference Phase**
**for** each test sample $(\boldsymbol{x}, y) \in \mathcal{D}_{\text{test}}$ **do**
    $\tau \leftarrow$ OOD_Model.predict($\boldsymbol{x}$) ▷ Compute OOD score
    **if** $\tau < \tau_{\text{threshold}}$ **then**
        $y_{\text{pred}} \leftarrow$ Prediction_Model.predict($\boldsymbol{x}$)
        **Output:** $y_{\text{pred}}$
    **else**
        **Output:** Reject sample $\boldsymbol{x}$
    **end if**
**end for**

**Fig. 8 Pseudocode of ODROP Framework**

The pseudocode of the ODROP method is shown in Fig. 8 (Algorithm 1). In the ODROP method, an OOD detection model and a prediction model are first trained. Subsequently, for each test data point, an OOD score is calculated using the OOD detection model. If this score is below a predefined threshold, predictions are made using the prediction model. Further details on the OOD score and the prediction model are discussed in the following sections.

### OOD detection model

Machine learning models assume that the test data come from the same distribution as the training data and may not perform accurately on OOD test data that deviate from the training data distribution. Identifying OOD data is crucial and is referred to as OOD detection[16,18]. OOD detection models compute an OOD score indicating the "likelihood" that the input data is OOD. Each input datum is classified as ID if the OOD score is below a certain threshold and OOD otherwise.

OOD detection models have evolved considerably and are categorized into generative and classification model-based approaches[16,18]. Traditionally, these models are benchmarked using existing image databases and manually separated into ID and OOD datasets to assess the binary classification performance (OOD-AUROC, OOD-PRAUC) [17,20]. Recently, classification-model-based approaches have been proposed in the image domain[29,30,36], building on the foundations established by generative model-based methods[37,38], reflecting advancements in accurately identifying OOD data. However, tabular data requires advanced domain knowledge of experts to distinguish ID and OOD datasets, and they have not been benchmarked, particularly health and medical data. In this study, we employed representative OOD detection methods such as the generative model-based VAE[23,27], the neural network classification model-based ensemble method[28], and GEM[30], a method developed based on neural network energy[29], recently developed and proposed in the field of imaging as an OOD detection model. Table 2 lists the name of each OOD detection method, its OOD score, and the calculation method.

The definitions of each OOD score (VAE reconstruction loss, ensemble std, ensemble epistemic, energy, and GEM scores) are as follows:

**VAE Reconstruction Loss (VAE reconstruction) Score**

$$\mathrm{Score}_{\mathrm{Reconstruction}}(\boldsymbol{x}) \triangleq E_{z\sim q^{\mathrm{VAE}}(\boldsymbol{z}|\boldsymbol{x})}\left[\|\boldsymbol{x} - f_{\mathrm{VAE}}(\boldsymbol{z})\|^2\right] \quad (1)$$

where $\boldsymbol{x}$ is the $m$-dimensional input vector. ($q^{\mathrm{VAE}}$, $f_{\mathrm{VAE}}$) are the encoder and decoder model obtained using VAE, respectively. The score is calculated by taking the expected value after 10-samplings according to the encoder model $q^{\mathrm{VAE}}$.

**Ensemble Prediction Probability Standard Deviation (ensemble std) Score**

$$\mathrm{Score}_{\mathrm{std}}(\boldsymbol{x}) \triangleq \sqrt{\frac{1}{M}\sum_{i=1}^{M}(p_i(\boldsymbol{x}) - p_{\mathrm{ensemble}}(\boldsymbol{x}))^2} \quad (2)$$

$$p_{\mathrm{ensemble}}(\boldsymbol{x}) \triangleq \frac{1}{M}\sum_{i=1}^{M} p_i(\boldsymbol{x}) \quad (3)$$

where $M$ is the number of neural network ensemble models and $p_i(\boldsymbol{x})$ is the prediction probability when $\boldsymbol{x}$ is the $m$-dimensional input vector.

**Ensemble Epistemic Uncertainty (ensemble epistemic) Score**

$$\mathrm{Score}_{\mathrm{epistemic}}(\boldsymbol{x}) \triangleq u_{\mathrm{total}}(\boldsymbol{x}) - u_{\mathrm{aleatoric}}(\boldsymbol{x}) \quad (4)$$

$$u_{\mathrm{total}}(\boldsymbol{x}) \triangleq -\sum_{y\in Y}\left(\frac{1}{M}\sum_{i=1}^{M} p(y|f_i,\boldsymbol{x})\right)\log_2\left(\frac{1}{M}\sum_{i=1}^{M} p(y|f_i,\boldsymbol{x})\right) \quad (5)$$

$$u_{\mathrm{aleatoric}}(\boldsymbol{x}) \triangleq -\frac{1}{M}\sum_{i=1}^{M}\sum_{y\in Y} p(y|f_i,\boldsymbol{x})\log_2 p(y|f_i,\boldsymbol{x}) \quad (6)$$

where $\boldsymbol{x}$ is an $m$-dimensional input feature vector, $Y$ is the label space, $M$ is the number of neural network ensemble models, and $f_i$ represents each ensemble model.

**Energy Score**

The Helmholtz free energy in deep neural networks is given as follows:

$$\mathrm{Score}_{\mathrm{Energy}}(\boldsymbol{x}) \triangleq -T\log\left(\sum_{j=1}^{K}\exp\left(\frac{f_j(\boldsymbol{x})}{T}\right)\right) \quad (7)$$

where $\boldsymbol{x}$ it the $m$-dimensional input feature vector, $T$ is the temperature parameter, and $K$ is the number of maximum classes. This Energy Score can be calculated easily using the LogSumExp operator. In this case, $K = 2$, because we used it for binary classification. In addition, $T = 1$ was used.

**GEM (Gaussian mixture based Energy Measurement) Score**

$$\mathrm{Score}_{\mathrm{GEM}}(\boldsymbol{x}) \triangleq -\log\left(\sum_{j=1}^{K}\exp\left(f_j(\boldsymbol{x}|\boldsymbol{\theta})\right)\right) \tag{8}$$

where $\boldsymbol{x}$ is the *m*-dimensional input feature vector.

$$f_j(\boldsymbol{x}|\boldsymbol{\theta}) \triangleq -\frac{1}{2}\left(h(\boldsymbol{x}|\boldsymbol{\theta}) - \hat{\boldsymbol{\mu}}_j\right)^{\top}\hat{\boldsymbol{\Sigma}}^{-1}\left(h(\boldsymbol{x}|\boldsymbol{\theta}) - \hat{\boldsymbol{\mu}}_j\right) \tag{9}$$

$$\hat{\boldsymbol{\mu}}_j \triangleq \frac{1}{|\mathcal{I}_j|}\sum_{i\in\mathcal{I}_j} h(\boldsymbol{x}_i|\boldsymbol{\theta}) \tag{10}$$

$$\hat{\boldsymbol{\Sigma}} \triangleq \frac{1}{\sum_{j=1}^{K}|\mathcal{I}_j|}\sum_{j=1}^{K}\sum_{i\in\mathcal{I}_j}\left(h(\boldsymbol{x}_i|\boldsymbol{\theta}) - \hat{\boldsymbol{\mu}}_j\right)\left(h(\boldsymbol{x}_i|\boldsymbol{\theta}) - \hat{\boldsymbol{\mu}}_j\right)^{\top} \tag{11}$$

where $h(\boldsymbol{x};\theta)$ is the m-dimensional output feature vector calculated using neural network model *f*. We assume that this feature vector space follows a *K*-class conditional multivariate Gaussian distribution. The distribution parameters $(\hat{\boldsymbol{\mu}}_j, \hat{\boldsymbol{\Sigma}})$ for each class are defined by the training data. $\mathcal{I}_j$ is the index set belonging to class *j* in the training data, and $|\mathcal{I}_j|$ is the number of elements.

We used all 334 features from the Hirosaki health checkup data to train the OOD detection models. The VAE model had a hidden layer size of 200, latent dimension of 75, learning rate of 1e-03, and maximum epoch of 400. The hidden layers of the NN Classification model were 200 and 50, batch size was 32, learning rate was 1e-03, maximum epoch was 100, and disease onset labels within a year were the target variables. For the ensemble method, five NN Classification models were trained using different seed values.

**Development of Disease Onset Prediction Models within one year**

**Disease Onset within one year Labels**

Diabetes, hypertension, and dyslipidemia were selected as lifestyle-related diseases. We assigned '1' for individuals diagnosed with the specified disease within one year from the measurement year and '0' otherwise. Diagnostic criteria[39–42] for determining disease onset were based on specific medical standards, as listed in Supplementary Table 2. Data with missing items were excluded to ensure accurate labeling of disease onset (Fig. 7).

**Training of Disease Onset Prediction Model**

We used Hirosaki health checkup data as the training data and developed three binary classification models using XGBoost[43] for each disease onset prediction model within a year. We performed feature selection using recursive feature elimination[44] and narrowed down all 334 features to the most relevant 20 features, given in Supplementary Table 3, for each model, XGBoost parameters were optimized using a grid search, as shown in Supplementary Table 4. The selected parameters are same for all disease prediction model— n_estimators: 6, min_child_weight: 1, and max depth: 6.

**Evaluation of OOD detection models in ODROP method**

OOD detection models calculate OOD scores, which indicate the extent to which data are OOD. Scores below a threshold are classified as ID, and those above as OOD. We used a rejection rate metric to evaluate the OOD detection model independent of the OOD score threshold. This rejection rate metric R measures the proportion of rejected test data (excluded from the prediction) based on the OOD score.

$$R = \frac{N_{\mathrm{OOD}}}{N_{\mathrm{ID}} + N_{\mathrm{OOD}}} \tag{12}$$

First, we varied the OOD score threshold to gradually reduce it. We then constructed a rejection curve[31] by plotting the rejection rate at each OOD score threshold on the horizontal axis and the corresponding prediction performance metric on the vertical axis. An upward trend in the rejection curve indicates improved prediction performance metrics for the test data, including the dataset shift. In this study, we used the AUROC and PRAUC as predictive performance metrics to conduct a qualitative evaluation of the most effective OOD detection model based on the improvement range and rejection rate at the maximum improvement observed in the rejection curve. We applied this approach to predict the onset of diabetes, hypertension, and dyslipidemia within one year. Additionally, we quantitatively assessed the rank correlation coefficient between the rejection rate and performance metrics by employing Kendall's tau rank correlation coefficient to evaluate the performance improvement stability by increasing the rejection rate. A positive coefficient indicates a progressive improvement in predictive performance with increasing rejection rate; higher values suggest a more stable improvement. We used a maximum rejection rate of 40% to calculate the rejection curve and rank the correlation coefficient.

### Discovery of Dataset Shift for Disease Onset Prediction Model

To identify important dataset shift items for the disease onset prediction model, we conducted hierarchical clustering using SHAP[24,45], highlighting the contribution of each item in the prediction model. Hierarchical clustering was applied to the Wakayama health checkup data, in which each disease occurred within one year, using the Ward aggregation and Euclidean distance. We then created ID and OOD data labels using the OOD score at the rejection rate, considering the maximum improvement in the AUROC-rejection curve as the threshold.

### Data Availability

The health checkup data used were collected from the Iwaki Health Promotion Project and the Wakayama study and were anonymized, and transferred to a secure data center with access restrictions. Anonymized data are available only to researchers for academic purposes who meet the access criteria provided by the Hirosaki University Faculty of Medicine (e-mail: coi@hirosaki-u.ac.jp), which requires approval from the ethics review committees of the Hirosaki University Faculty of Medicine and the researcher's affiliated institutions. Additional data are available upon reasonable request from the corresponding author.

## Acknowledgements

This research was supported by the JST COI Program (JPMJCE1302), the JST COI-NEXT program (JPMJCA2201), and the 2023 Iwadare Scholarship Association Research Grant.

## Author Contributions

Hirosaki health checkup data collection: Y.T., T.M., K.M., and S.N. Wakayama health checkup data collection: M.A. and N.M. Original concept, experiments conduction, and figures preparation: T.T. Manuscript writing and revising: T.T., E.U., R.K., Y.M., Y.O, M.A., N.M., Y.T., T.M., K.M., S.N., and Y.O. All authors contributed to manuscript preparation and approved the publication of it.

## Additional Information

## Competing Interests

The authors declare no competing interests.

### Supplementary Information

The supplementary material is available.

### Code Availability

The code for OOD detection in tabular data we used includes https://github.com/clinfo/OOD4Tab.

## Tables

**Table 1. Summary Statistics (Mean ± std) and Test p-values for Main Items in Hirosaki and Wakayama Health Checkups**

| Items | Hirosaki data | Wakayama data | p-value |
|---|---|---|---|
| Age [year] | 55.6 ± 14.9 | 65.3 ± 10.8 | 1.5e-18 |
| Gender | Male: 5975 (38.7%)<br>Female: 9479(61.3%) | Male: 672 (43.9%)<br>Female: 859(56.1%) | 7.1e-05 |
| Height [cm] | 159.7 ± 9.2 | 160.0 ± 9.0 | 0.23 |
| Body Weight [kg] | 58.9 ± 11.3 | 59.0 ± 11.6 | 0.63 |
| BMI [kg/m$^2$] | 23.0 ± 3.3 | 22.9 ± 3.4 | 0.53 |
| Serum Glucose [mg/dL] | 90.3 ± 18.1 | 95.7 ± 18.6 | 2.7e-26 |
| HbA1c (NGSP method) [%] | 5.7 ± 0.6 | 5.8 ± 0.5 | 1.9e-14 |
| Estimated glomerular filtration rate (eGFR) | 79.9 ± 15.7 | 66.3 ± 12.6 | 5.5e-25 |
| Systolic Blood Pressure [mmHg] | 126.5 ± 18.9 | 129.1 ± 19.1 | 3.8e-07 |
| Diastolic Blood Pressure [mmHg] | 75.4 ± 11.9 | 78.6 ± 11.2 | 1.8e-26 |
| HDL Cholesterol [mg/dL] | 64.3 ± 16.5 | 64.2 ± 17.5 | 0.71 |
| LDL Cholesterol [mg/dL] | 116.3 ± 28.9 | 121.7 ± 32.5 | 5.6e-10 |
| Total Cholesterol [mg/dL] | 98.1 ± 77.1 | 111.2 ± 82.4 | 2.7e-09 |
| RbaPWV [cm/s] | 1514 ± 379 | 1579 ± 379 | 2.2e-10 |
| LbaPWV [cm/s] | 1522 ± 376 | 1590 ± 392 | 1.0e-10 |
| Right_Arm-Body_Fat_Percentage_Score | 2.3 ± 10.5 | -0.3 ± 1.7 | 7.5e-16 |
| Diabetes Onset in one year | Onset: 258<br>No Onset: 10101 | Onset: 12<br>No Onset: 300 | 0.19 |
| Dyslipidemia Onset in one year | Onset: 1508<br>No Onset: 3093 | Onset: 22<br>No Onset: 71 | 0.081 |
| Hypertension Onset in one year | Onset: 1139<br>No Onset: 5510 | Onset: 33<br>No Onset: 146 | 0.72 |

Summary statistics are presented as the mean ± standard deviation. Welch's t-test was used for continuous variables, while $\chi^2$ test and Fisher's exact test were applied according to Cochran's rule for discrete variables.

**Table 2 OOD detection method**

| Based model for OOD detection | | OOD score name | Proposal is image-based |
|---|---|---|---|
| Variational auto-encoder (VAE) | Density-based | VAE reconstruction loss (VAE reconstruction) | No |
| Ensemble of neural network classification model (Ensemble) | Classification model-based | Ensemble prediction Probability Standard Deviation (ensemble std) | No |
| | | Epistemic uncertainty (ensemble epistemic) | No |
| Neural network classification model (NN) | | Energy Score | Yes |
| | | Gaussian mixture based Energy Measurement (GEM) Score | Yes |

# Supplementary Material: Out-of-distribution Reject Option Method for Dataset Shift Problem in Early Disease Onset Prediction

Taisei Tosaki[1], Eiichiro Uchino[1], Ryosuke Kojima[1], Yohei Mineharu[1], Yuji Okamoto[1], Mikio Arita[2], Nobuyuki Miyai[2] , Yoshinori Tamada[3], Tatsuya Mikami[4], Koichi Murashita[5], Shigeyuki Nakaji[4], Yasushi Okuno[1,*]

[1] Graduate School of Medicine, Kyoto University, Japan

[2] Graduate School of Health and Nursing Science, Wakayama Medical University, Japan

[3] Research Center for Health-Medical Data Science, Graduate School of Medicine, Hirosaki University, Japan

[4] Innovation Center for Health Promotion, Graduate School of Medicine, Hirosaki University, Japan

[5] Center of Innovation Research Initiatives Organization, Hirosaki University, Japan

[*] okuno.yasushi.4c@kyoto-u.ac.jp

# Supplementary Material: Out-of-distribution Reject Option Method for Dataset Shift Problem in Early Disease Onset Prediction

## Supplementary Note 1. Comparison of PRAUC Baselines between Hirosaki and Wakayama Health Checkups

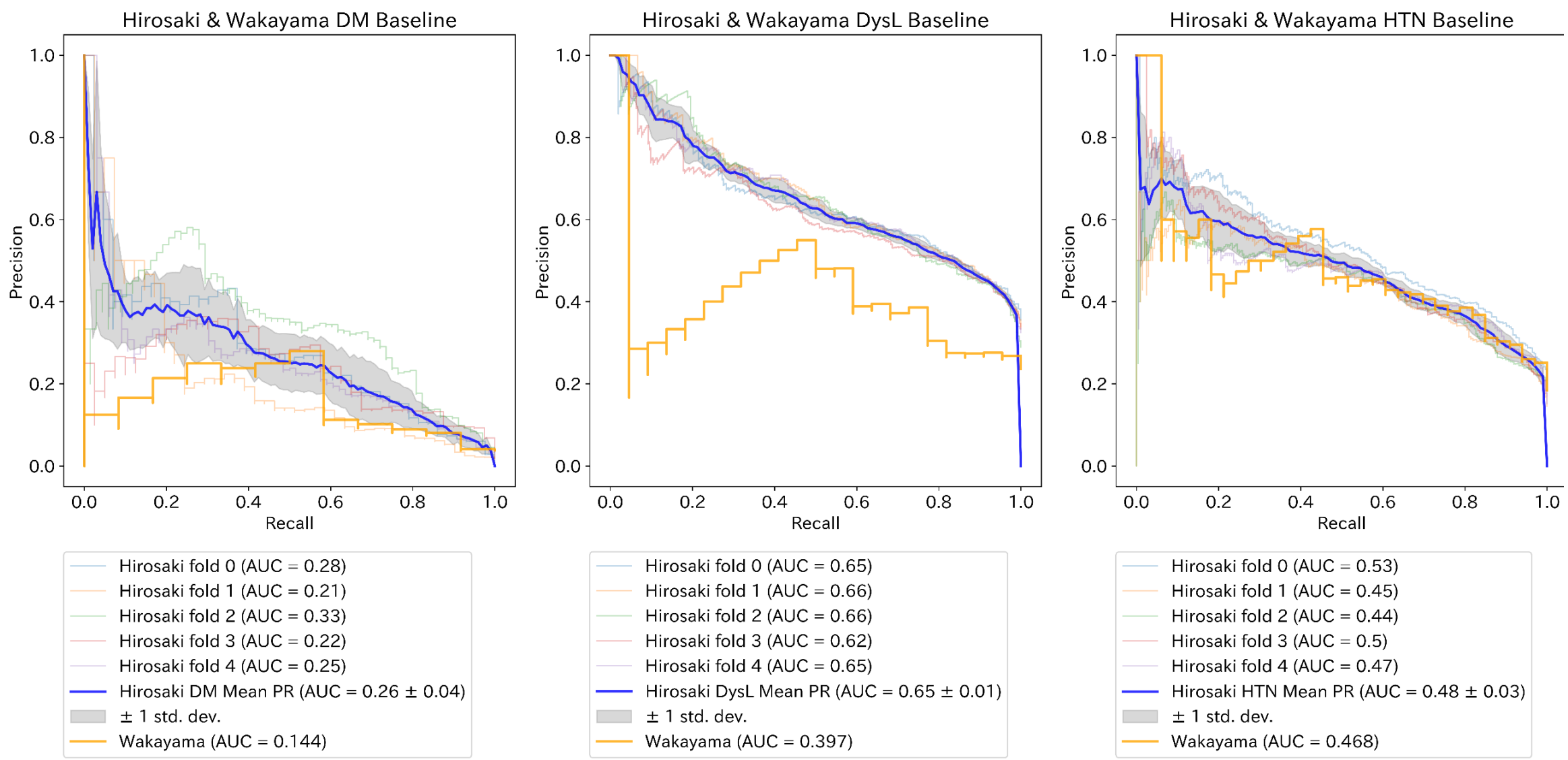

**Supplementary Fig. 1 Comparison of PRAUC Baselines between Hirosaki and Wakayama Health Checkups**

Results of the 5-fold cross-validation PR curves for each disease onset prediction task conducted in Hirosaki, along with their mean ± std PR curves, compared to the PR curve results from Wakayama health checkup data. The values in parentheses represent the PRAUC values.

(Left): Prediction of diabetes onset within one year

(Center): Prediction of dyslipidemia onset within one year

(Right): Prediction of hypertension onset within one year

## Supplementary Note 2. PRAUC-rejection rate rank correlation coefficients and PRAUC-rejection curves in Hirosaki Health Checkup data

A

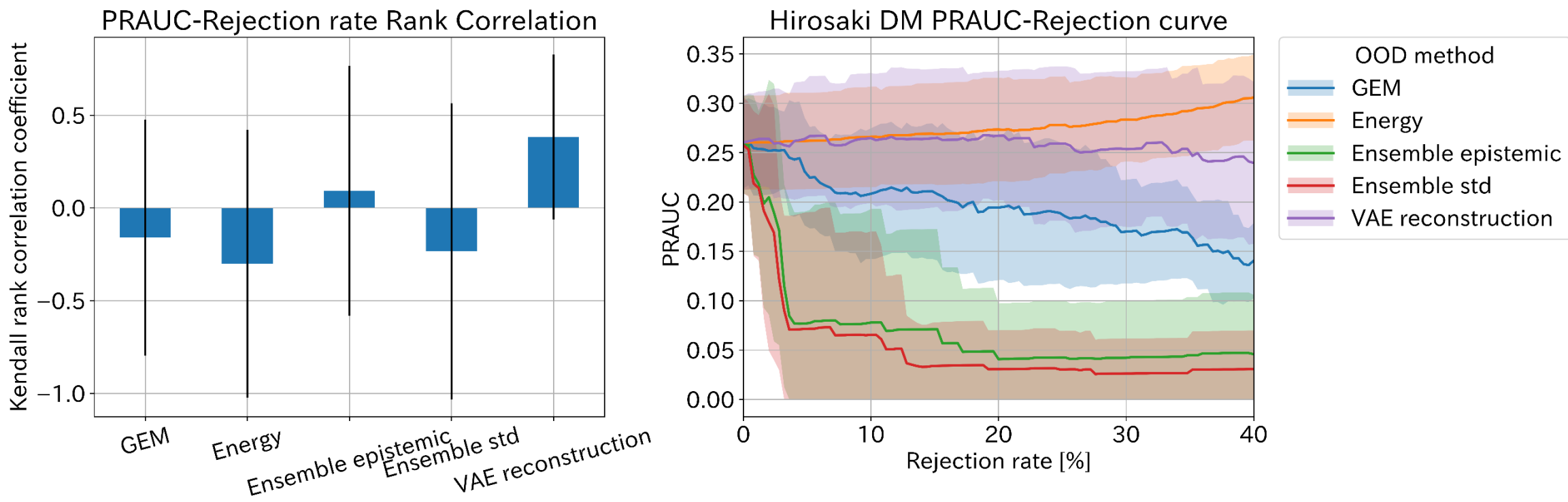

B

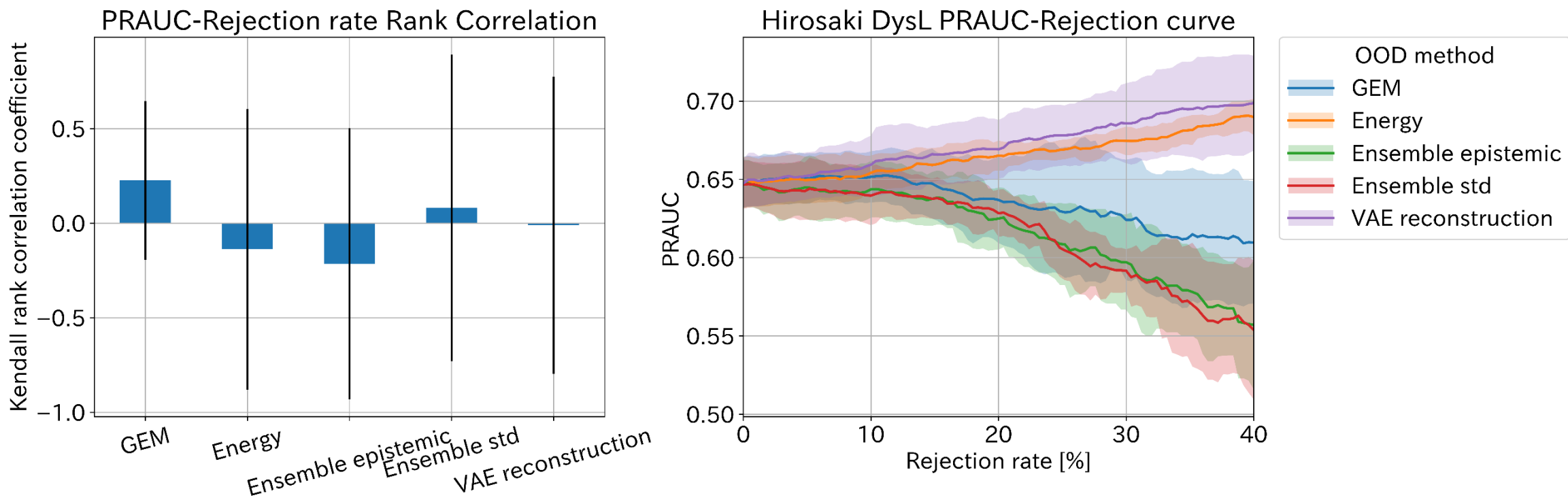

C

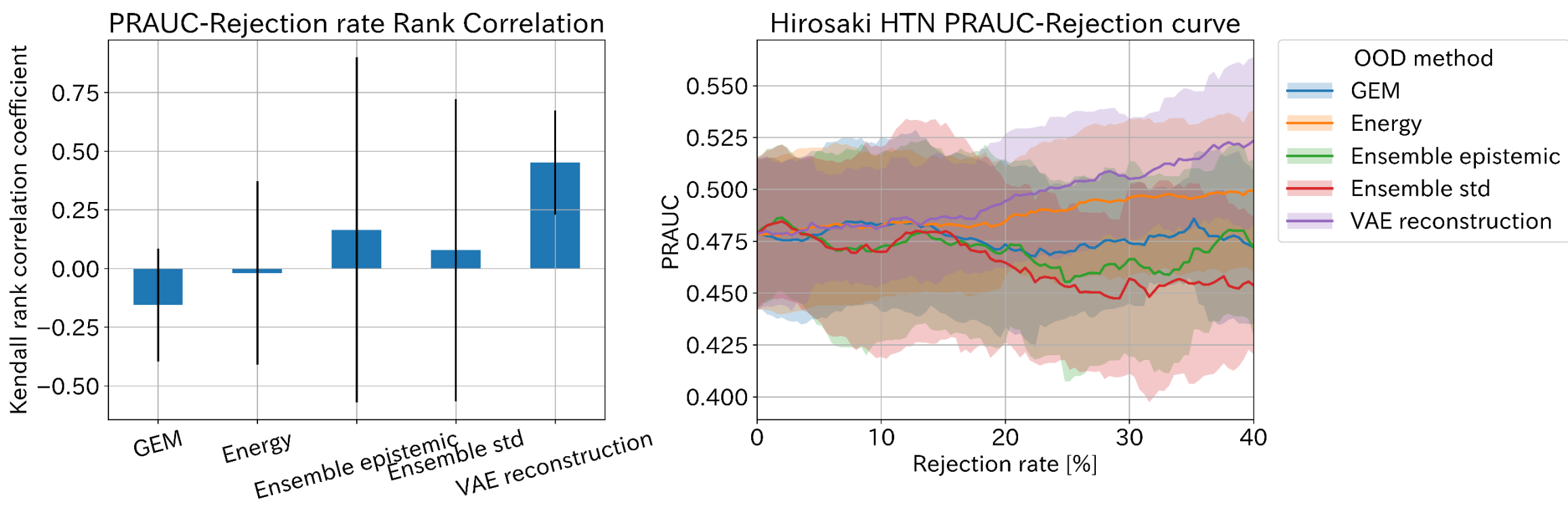

**Supplementary Fig. 2 PRAUC-rejection rate rank correlation coefficients and PRAUC-rejection curves in Hirosaki Health Checkup data**

**A: Diabetes Melius (DM), B: Dyslipidemia (DysL), C: Hypertension (HTN).**

Left Bar Plot: Mean ± std of rank correlation coefficient between rejection rate and PRAUC

Right: PRAUC-Rejection curve. Y-axis represents the PRAUC value (mean ± std) and x-axis is the rejection rate.

The VAE reconstruction method showed positive and substantial rank correlation coefficients for diabetes and hypertension (A and C), with the rejection curve indicating a near-monotonic improvement. The OOD detection method with the largest improvement from the baseline is VAE reconstruction for dyslipidemia and hypertension (B and C) and energy for diabetes (A), where VAE reconstruction maintained a performance equivalent to the baseline.

**Supplementary Note 3. PRAUC-rejection rate rank correlation coefficients and PRAUC-rejection curves in Wakayama Health Checkup data**

A

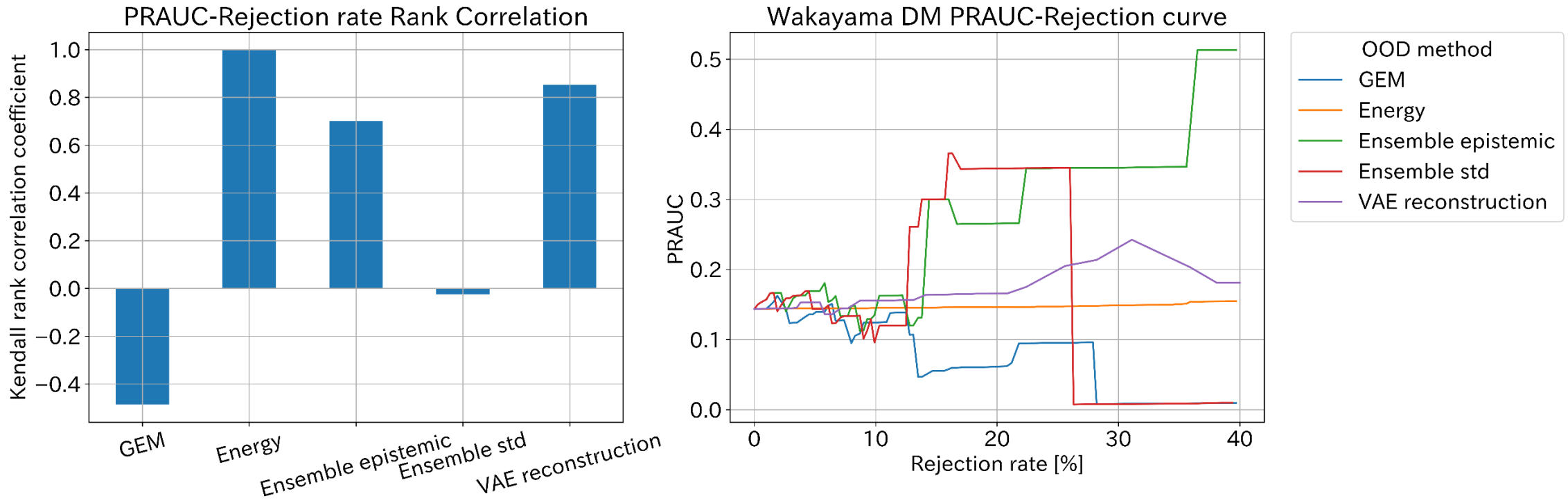

B

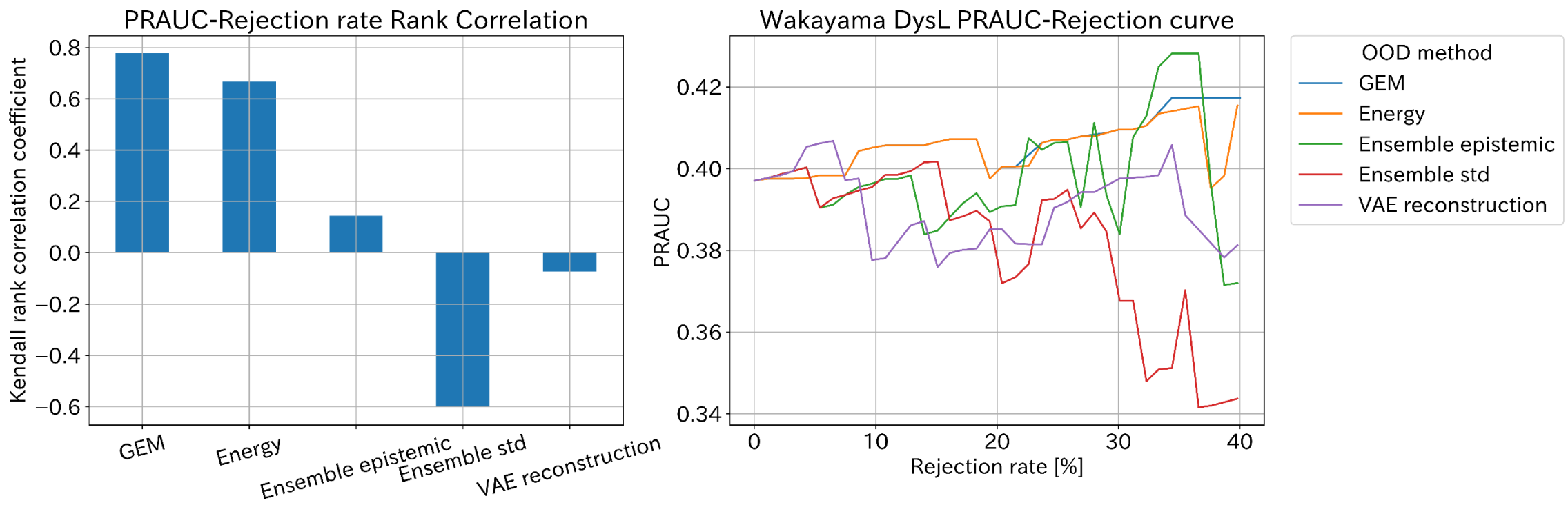

C

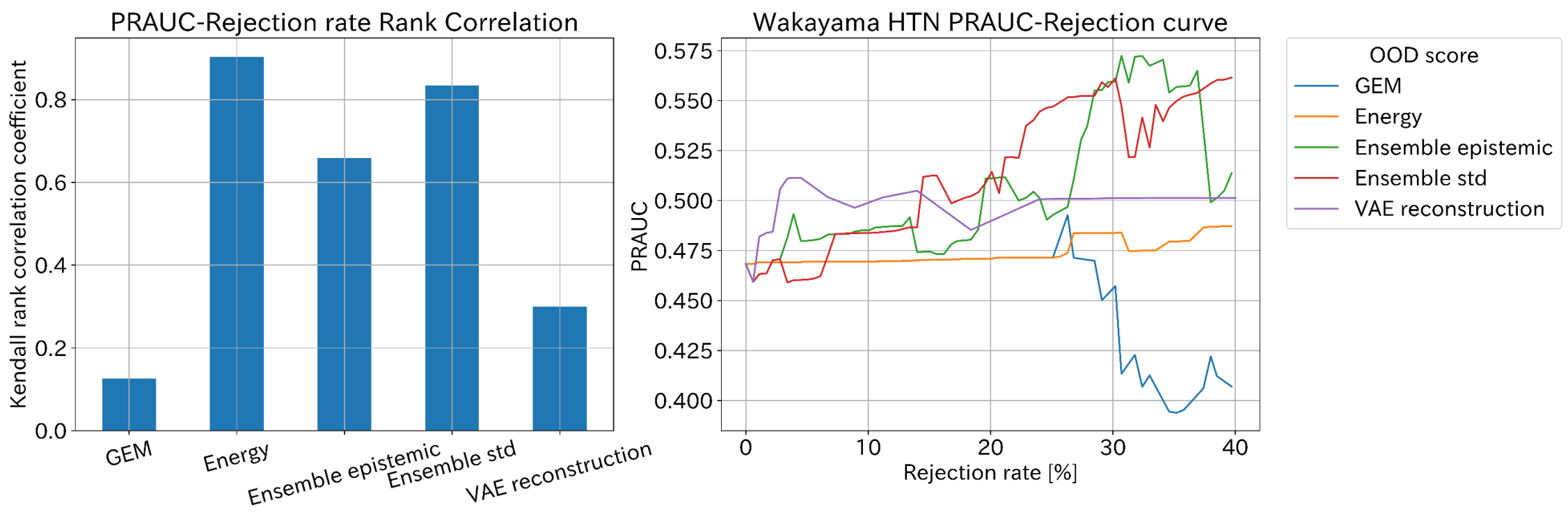

**Supplementary Fig. 3 PRAUC-Rejection rate rank correlation coefficients and PR-AUC rejection curves in Wakayama Health Checkup data**

**A: Diabetes Melius (DM), B: Dyslipidemia (DysL), C: Hypertension (HTN).**

Left Bar Plot: The rank correlation coefficient between rejection rate and PRAUC.

Right: PRAUC-Rejection curve. The Y- and X-axes represent the PRAUC value and rejection rate, respectively.

Energy and ensemble epistemic show positive rank correlation coefficients across all diseases (A, B, and C), with ensemble epistemic demonstrating the largest improvement from baseline. However, the performance of the ensemble epistemic method did not consistently maintain its peak improvement at higher rejection rates. In contrast, VAE reconstruction method, which did not always reach the highest improvement margins, maintains its peak performance better at the maximum improvement.

**Supplementary Note 4. SHAP Clustering for Dyslipidemia and Hypertension Onset within one-year records from Wakayama Health Checkups**

B

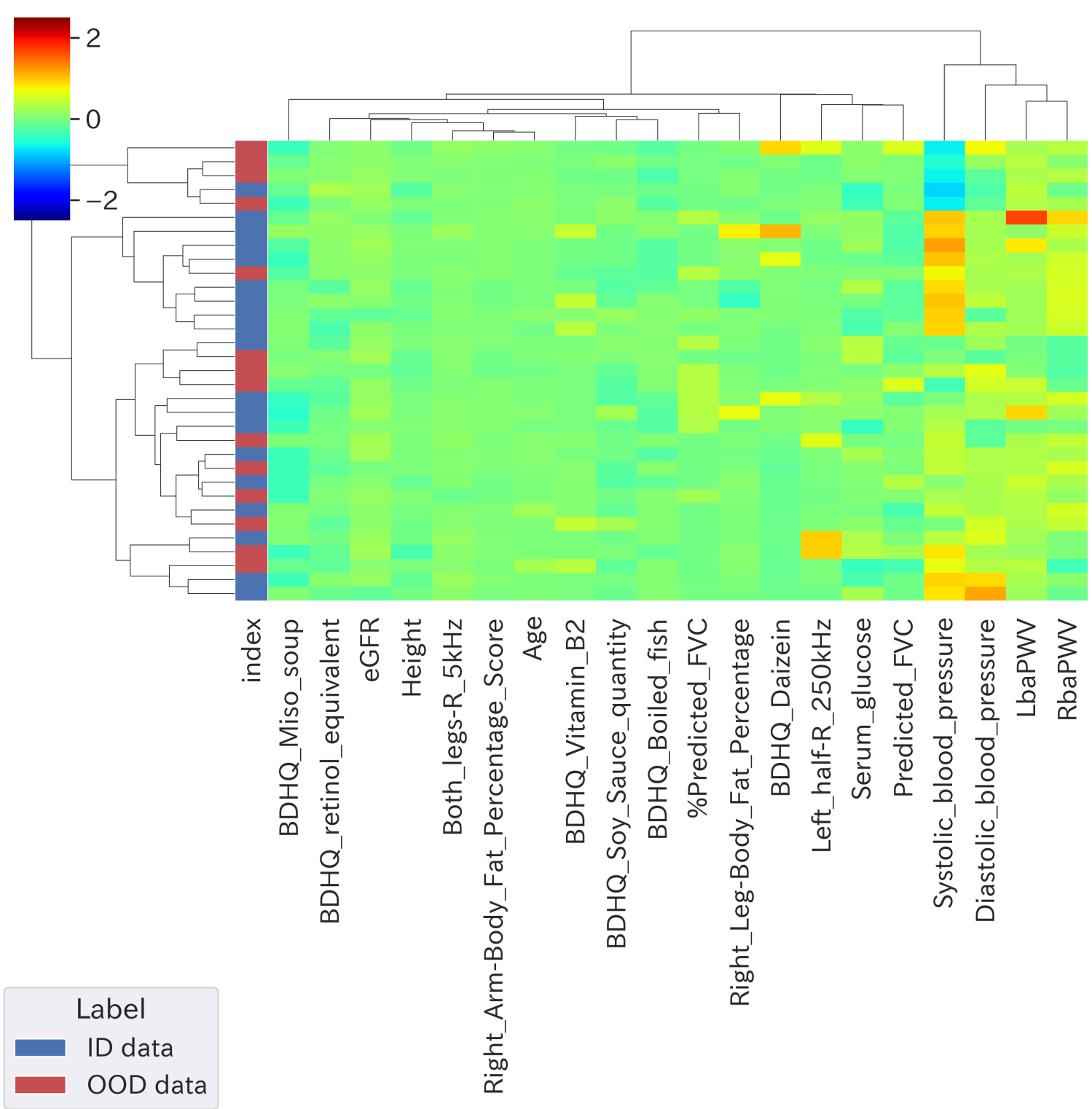

**Supplementary Fig. 4 B SHAP Clustering for Hypertension Onset within one-year records in Wakayama Health Checkups**

This figure shows a hierarchical clustering analysis using SHAP values from a 1-year hypertension onset prediction model for individuals from Wakayama health checkup data who developed diabetes within a year. A colormap represents the magnitude of the SHAP values calculated by the prediction model, with the vertical axis listing the Wakayama health checkup data of individuals who developed diabetes within a year. The horizontal axis without an index column shows the names of each examination item used in the prediction model, whereas an index column is IDs and OOD labels based on the Epistemic threshold at the rejection rate of 30.7%, where AUROC was maximized in the rejection curve.

Based on this clustering, when the Wakayama health checkup data was divided into two major groups, the group with lower SHAP values for systolic blood pressure could be considered part of the OOD data group.

## Supplementary Note 5. All Items of Characteristics List

**Supplementary Table 1. All Items Mean ± Std (Median / Rate) and p-value**

| Item | Cate-gory | Hirosaki Health Checkup | Wakayama Health Checkup | p-value |
|---|---|---|---|---|
| Sleep_Time_required_to_go_to_bed | | 14.2 ± 15.5 (10.0) | 18.9 ± 18.7 (10.0) | 2.9e-20 |
| Sleep_Inability_to_sleep_Sleeping_within_30_minutes | 'None' | 10806 (73.0%) | 812 (53.1%) | 4.1e-59 |
| Sleep_Inability_to_sleep_Sleeping_within_30_minutes | Less than once a week | 2123 (14.4%) | 353 (23.1%) | |
| Sleep_Inability_to_sleep_Sleeping_within_30_minutes | 1-2 times a week | 1110 (7.5%) | 223 (14.6%) | |
| Sleep_Inability_to_sleep_Sleeping_within_30_minutes | At least 3 times a week | 754 (5.1%) | 140 (9.2%) | |
| Sleep_unable_to_sleep_early_morning_wake_at_night | 'None' | 10078 (68.3%) | 399 (26.3%) | 4.0e-275 |
| Sleep_unable_to_sleep_early_morning_wake_at_night | Less than once a week | 2049 (13.9%) | 324 (21.4%) | |
| Sleep_unable_to_sleep_early_morning_wake_at_night | 1-2 times a week | 1516 (10.3%) | 367 (24.2%) | |
| Sleep_unable_to_sleep_early_morning_wake_at_night | At least 3 times a week | 1122 (7.6%) | 427 (28.1%) | |
| Sleep_Unable_to_sleep_Toilet | 'None' | 10061 (68.1%) | 355 (23.3%) | 0.0 |
| Sleep_Unable_to_sleep_Toilet | Less than once a week | 1935 (13.1%) | 298 (19.6%) | |
| Sleep_Unable_to_sleep_Toilet | 1-2 times a week | 1373 (9.3%) | 297 (19.5%) | |
| Sleep_Unable_to_sleep_Toilet | At least 3 times a week | 1401 (9.5%) | 573 (37.6%) | |
| Sleep_unable_to_sleep_difficult_to_breathe | At least 3 times a week | 34 (0.2%) | 10 (0.7%) | 7.7e-45 |
| Sleep_unable_to_sleep_difficult_to_breathe | 1-2 times a week | 92 (0.6%) | 37 (2.4%) | |
| Sleep_unable_to_sleep_difficult_to_breathe | 'None' | 14380 (97.3%) | 1377 (90.4%) | |
| Sleep_unable_to_sleep_difficult_to_breathe | Less than once a week | 272 (1.8%) | 100 (6.6%) | |
| Sleep_unable_to_sleep_cough_snore | 'None' | 13709 (92.8%) | 1072 (70.6%) | 9.8e-186 |
| Sleep_unable_to_sleep_cough_snore | Less than once a week | 614 (4.2%) | 203 (13.4%) | |
| Sleep_unable_to_sleep_cough_snore | 1-2 times a week | 275 (1.9%) | 127 (8.4%) | |
| Sleep_unable_to_sleep_cough_snore | At least 3 times a week | 179 (1.2%) | 116 (7.6%) | |
| Sleep_unable_to_sleep_cold | 'None' | 13864 (93.8%) | 1417 (93.0%) | 0.054 |
| Sleep_unable_to_sleep_cold | Less than once a week | 675 (4.6%) | 77 (5.1%) | |
| Sleep_unable_to_sleep_cold | 1-2 times a week | 200 (1.4%) | 20 (1.3%) | |
| Sleep_unable_to_sleep_cold | At least 3 times a week | 34 (0.2%) | 9 (0.6%) | |
| Sleep_Unable_to_sleep_Heat | At least 3 times a week | 70 (0.5%) | 52 (3.4%) | 1.7e-195 |
| Sleep_Unable_to_sleep_Heat | 'None' | 13458 (91.1%) | 1017 (66.7%) | |
| Sleep_Unable_to_sleep_Heat | Less than once a week | 968 (6.5%) | 302 (19.8%) | |
| Sleep_Unable_to_sleep_Heat | 1-2 times a week | 284 (1.9%) | 153 (10.0%) | |
| Sleep_not_able_to_sleep_nightmares | Less than once a week | 832 (5.6%) | 167 (11.0%) | 3.3e-20 |
| Sleep_not_able_to_sleep_nightmares | At least 3 times a week | 69 (0.5%) | 8 (0.5%) | |
| Sleep_not_able_to_sleep_nightmares | 1-2 times a week | 204 (1.4%) | 45 (3.0%) | |
| Sleep_not_able_to_sleep_nightmares | 'None' | 13667 (92.5%) | 1303 (85.6%) | |
| Sleep_unable_to_sleep_pain | 'None' | 13490 (91.4%) | 1166 (76.7%) | 2.3e-73 |
| Sleep_unable_to_sleep_pain | Less than once a week | 635 (4.3%) | 197 (13.0%) | |
| Sleep_unable_to_sleep_pain | 1-2 times a week | 387 (2.6%) | 104 (6.8%) | |
| Sleep_unable_to_sleep_pain | At least 3 times a week | 252 (1.7%) | 53 (3.5%) | |
| Sleep_Sleeping_Medication_Frequency | At least 3 times a week | 509 (3.4%) | 123 (8.0%) | 1.2e-19 |
| Sleep_Sleeping_Medication_Frequency | 'None' | 13947 (94.4%) | 1352 (88.5%) | |
| Sleep_Sleeping_Medication_Frequency | Less than once a week | 199 (1.3%) | 35 (2.3%) | |
| Sleep_Sleeping_Medication_Frequency | 1-2 times a week | 125 (0.8%) | 18 (1.2%) | |
| Sleep_Drowsiness_At_Work | 1-2 times a week | 609 (4.1%) | 43 (2.8%) | 7.7e-06 |
| Sleep_Drowsiness_At_Work | At least 3 times a week | 267 (1.8%) | 18 (1.2%) | |
| Sleep_Drowsiness_At_Work | 'None' | 12673 (85.8%) | 1289 (84.4%) | |
| Sleep_Drowsiness_At_Work | Less than once a week | 1230 (8.3%) | 177 (11.6%) | |
| Torso-lean_mass | | 22.3 ± 7.4 (22.2) | 24.4 ± 4.3 (23.2) | 9.4e-57 |
| Gender | Male | 5975 (38.7%) | 672 (43.9%) | 7.1e-05 |
| Gender | Female | 9479 (61.3%) | 859 (56.1%) | |
| Age | | 55.6 ± 14.9 (58.0) | 65.3 ± 10.8 (67.0) | 1.5e-188 |
| Height | | 159.7 ± 9.2 (159.0) | 160.0 ± 9.0 (159.7) | 0.23 |
| body weight | | 58.9 ± 11.3 (57.4) | 59.0 ± 11.6 (57.6) | 0.63 |
| BMI | | 23.0 ± 3.3 (22.7) | 22.9 ± 3.4 (22.6) | 0.53 |
| Systolic_blood_pressure | | 126.5 ± 18.9 (125.0) | 129.1 ± 19.1 (127.0) | 3.8e-07 |
| Diastolic_blood_pressure | | 75.4 ± 11.9 (75.0) | 78.6 ± 11.2 (78.0) | 1.8e-26 |
| Limb_blood_pressure_LbaPWV | | 1522.0 ± 376.4 (1462.0) | 1590.3 ± 391.6 (1532.5) | 1.0e-10 |
| Limb_blood_pressure_RbaPWV | | 1514.3 ± 378.5 (1453.0) | 1579.3 ± 378.6 (1523.0) | 2.2e-10 |
| Uric_acid | | 5.0 ± 1.3 (4.8) | 5.3 ± 1.3 (5.2) | 3.7e-25 |
| Total_cholesterol | | 204.2 ± 33.7 (203.0) | 206.5 ± 38.0 (205.0) | 0.019 |
| HDL_cholesterol | | 64.3 ± 16.5 (63.0) | 64.2 ± 17.5 (62.0) | 0.71 |
| TG_triglycerides | | 98.1 ± 77.1 (79.0) | 111.2 ± 82.4 (93.0) | 2.7e-09 |
| Sodium(Na) | | 141.6 ± 1.8 (142.0) | 142.4 ± 1.9 (142.0) | 3.6e-52 |
| Potassium(K) | | 4.1 ± 0.4 (4.0) | 4.0 ± 0.3 (4.0) | 1.3e-10 |
| Chlorine(Cl) | | 103.8 ± 2.1 (104.0) | 105.1 ± 2.1 (105.0) | 1.3e-105 |
| Serum_iron | | 101.0 ± 37.3 (99.0) | 106.3 ± 33.6 (104.0) | 8.6e-09 |
| Right_Arm-Body_Fat_Percentage_Score | | 2.3 ± 10.5 (0.0) | -0.3 ± 1.7 (0.0) | 7.5e-159 |
| White_blood_cell_count | | 5231.6 ± 1516.2 (5000.0) | 5492.6 ± 1456.7 (5300.0) | 3.3e-11 |
| Red_blood_cell_count | | 454.9 ± 42.7 (453.0) | 450.5 ± 42.8 (448.0) | 0.00011 |
| Hemoglobin | | 13.8 ± 1.5 (13.8) | 13.8 ± 1.4 (13.7) | 0.31 |
| Hematocrit | | 43.6 ± 4.0 (43.5) | 41.9 ± 3.7 (41.7) | 1.6e-62 |
| MCV | | 96.1 ± 5.5 (96.0) | 93.2 ± 5.0 (93.0) | 3.9e-90 |
| MCH | | 30.5 ± 2.1 (30.6) | 30.7 ± 1.9 (30.7) | 1.1e-05 |
| MCHC%% (MCHC%) | | 31.7 ± 1.1 (31.7) | 32.9 ± 0.9 (32.9) | 0.0 |
| Platelet_count | | 23.9 ± 5.6 (23.3) | 24.8 ± 6.2 (24.2) | 3.3e-07 |
| Serum_glucose | | 90.3 ± 18.1 (87.0) | 95.7 ± 18.6 (91.0) | 2.7e-26 |
| Total_bilirubin | | 0.8 ± 0.3 (0.8) | 0.9 ± 0.3 (0.8) | 2.2e-13 |
| AST_GOT | | 23.1 ± 12.8 (21.0) | 22.5 ± 8.3 (21.0) | 0.0066 |
| ALT_GPT | | 21.8 ± 16.0 (18.0) | 20.4 ± 13.1 (17.0) | 7.6e-05 |
| gamma-GTP | | 32.8 ± 46.6 (21.0) | 32.9 ± 38.4 (22.0) | 0.92 |
| Total_protein | | 7.3 ± 0.4 (7.3) | 7.3 ± 0.4 (7.3) | 0.62 |
| Creatinine | | 0.7 ± 0.3 (0.7) | 0.8 ± 0.2 (0.8) | 6.5e-69 |
| Urea_nitrogen | | 14.7 ± 4.2 (14.1) | 16.1 ± 4.9 (15.0) | 6.4e-28 |
| Smoking_drinking_smoking_number_of_units | | 5.8 ± 10.3 (0.0) | 3.2 ± 7.8 (0.0) | 3.0e-19 |
| Sleep_Poor_quality_of_sleep | pretty bad | 1936 (14.1%) | 617 (40.6%) | 1.2e-218 |
| Sleep_Poor_quality_of_sleep | very bad | 180 (1.3%) | 102 (6.7%) | |
| Sleep_Poor_quality_of_sleep | Very good | 3256 (23.7%) | 111 (7.3%) | |
| Sleep_Poor_quality_of_sleep | Pretty good. | 8343 (60.8%) | 690 (45.4%) | |
| Daily_life_Good_health_status | Best for. | 241 (1.8%) | 32 (2.1%) | 1.3e-21 |
| Daily_life_Good_health_status | Very good. | 3078 (22.5%) | 252 (16.5%) | |
| Daily_life_Good_health_status | good | 7990 (58.4%) | 874 (57.2%) | |
| Daily_life_Good_health_status | Not so good. | 2021 (14.8%) | 327 (21.4%) | |
| Daily_life_Good_health_status | not good | 342 (2.5%) | 38 (2.5%) | |
| Daily_life_Good_health_status | Not good at all. | 0 (0.0%) | 5 (0.3%) | |
| Daily_Life_Body_Pain | light pain | 4316 (31.6%) | 460 (30.1%) | 4.2e-21 |
| Daily_Life_Body_Pain | Very severe pain | 92 (0.7%) | 7 (0.5%) | |
| Daily_Life_Body_Pain | severe pain | 765 (5.6%) | 59 (3.9%) | |
| Daily_Life_Body_Pain | It wasn't there at all. | 4017 (29.4%) | 374 (24.5%) | |
| Daily_Life_Body_Pain | faint pain | 2457 (18.0%) | 435 (28.5%) | |
| Daily_Life_Body_Pain | Moderate pain | 2020 (14.8%) | 192 (12.6%) | |
| Left_Leg-Body_Fat_Percentage | | 27.8 ± 7.4 (28.6) | 26.5 ± 7.8 (26.9) | 1.8e-10 |
| Left_leg-fat_content | | 2.9 ± 1.0 (2.8) | 2.7 ± 1.0 (2.6) | 4.2e-08 |
| Left_leg-lean_mass | | 7.5 ± 1.8 (7.0) | 7.6 ± 1.8 (7.1) | 0.27 |
| Left_leg-muscle_mass | | 7.1 ± 1.7 (6.6) | 7.1 ± 1.7 (6.7) | 0.26 |
| Left_Leg-Body_Fat_Percentage_Score | | 0.1 ± 1.4 (0.0) | -0.0 ± 1.5 (0.0) | 2.7e-05 |
| Left_Leg-Muscle_Mass_Score | | -0.7 ± 1.4 (-1.0) | -0.7 ± 1.4 (-1.0) | 0.27 |
| Right_Arm-Body_Fat_Percentage | | 21.4 ± 8.9 (20.2) | 20.9 ± 8.9 (19.6) | 0.047 |
| Right_arm-fat_content | | 0.6 ± 0.3 (0.6) | 0.6 ± 0.3 (0.5) | 0.27 |
| Right_arm-lean_mass | | 2.2 ± 0.7 (2.0) | 2.3 ± 0.6 (2.1) | 0.48 |
| Right_arm-muscle_mass | | 2.1 ± 0.6 (1.9) | 2.1 ± 0.6 (2.0) | 0.47 |
| Right_arm-muscle_mass_score | | 0.7 ± 1.4 (1.0) | 0.6 ± 1.4 (0.0) | 0.0055 |
| Left_Arm-Body_Fat_Percentage | | 22.6 ± 9.2 (21.6) | 22.2 ± 9.1 (20.9) | 0.064 |
| Left_arm-fat_content | | 0.6 ± 0.4 (0.6) | 0.6 ± 0.4 (0.6) | 0.35 |
| Left_arm-lean_mass | | 2.2 ± 0.6 (1.9) | 2.2 ± 0.6 (2.0) | 0.40 |
| Left_arm-muscle_mass | | 2.0 ± 0.6 (1.8) | 2.0 ± 0.6 (1.9) | 0.50 |
| Left_arm-Body_Fat_Percentage_Score | | -0.1 ± 1.7 (0.0) | -0.0 ± 1.7 (0.0) | 0.049 |
| Left_arm-muscle_mass_score | | 0.3 ± 1.4 (0.0) | 0.2 ± 1.4 (0.0) | 0.0020 |
| Torso-Body_Fat_Percentage | | 25.3 ± 9.0 (24.8) | 24.6 ± 9.1 (24.2) | 0.0029 |
| Torso-fat_mass | | 8.4 ± 3.9 (8.0) | 8.2 ± 4.0 (7.9) | 0.14 |
| Torso-Muscle_mass | | 22.8 ± 4.2 (21.3) | 23.1 ± 4.2 (21.8) | 0.0015 |
| Torso-Body_Fat_Percentage_Score | | -0.1 ± 1.5 (0.0) | -0.1 ± 1.5 (0.0) | 0.35 |
| Torso-Muscle_Mass_Score | | 0.6 ± 1.5 (1.0) | 0.7 ± 1.3 (1.0) | 0.0091 |
| Left_half-R_5kHz | | 691.1 ± 90.6 (683.9) | 656.7 ± 82.9 (652.2) | 1.4e-49 |
| Left_half-X_5kHz | | -28.2 ± 15.7 (-26.7) | -25.0 ± 11.0 (-23.9) | 2.4e-24 |
| Left_half-R_50kHz | | 619.2 ± 85.4 | 594.0 ± 78.6 | 9.1e-31 |
| | | (613.7) | (591.9) | |
| Left_half-X_50kHz | | -59.2 ± 9.8 (-58.6) | -54.6 ± 8.1 (-54.5) | 3.5e-83 |
| Left_half-R_250kHz | | 557.6 ± 80.1 (552.6) | 537.5 ± 73.8 (535.0) | 7.5e-23 |
| Left_half-X_250kHz | | -58.4 ± 27.4 (-57.3) | -59.0 ± 9.0 (-58.2) | 0.091 |
| Left_half-R_500kHz | | 542.6 ± 78.7 (537.7) | 0.0 ± 0.0 (0.0) | 0.0 |
| Left_half-X_500kHz | | -66.0 ± 16.7 (-63.6) | 0.0 ± 0.0 (0.0) | 0.0 |
| Right_foot-R_5kHz | | 280.0 ± 37.1 (279.0) | 260.8 ± 35.7 (259.7) | 7.5e-80 |
| Right_foot-X_5kHz | | -9.6 ± 3.5 (-9.5) | -8.6 ± 2.3 (-8.3) | 7.0e-59 |
| Right_foot-R_50kHz | | 253.0 ± 33.5 (251.7) | 237.4 ± 32.6 (235.5) | 2.3e-65 |
| Right_foot-X_50kHz | | -21.5 ± 4.6 (-21.4) | -19.2 ± 4.0 (-19.2) | 1.1e-85 |
| Right_leg-R_250kHz | | 231.7 ± 31.1 (230.3) | 217.2 ± 30.0 (215.2) | 7.8e-66 |
| Right_foot-X_250kHz | | -18.9 ± 3.6 (-18.8) | -17.3 ± 2.9 (-17.3) | 2.4e-89 |
| Right_foot-R_500kHz | | 225.2 ± 30.3 (223.7) | 0.0 ± 0.0 (0.0) | 0.0 |
| Right_foot-X_500kHz | | -20.4 ± 4.3 (-20.0) | 0.0 ± 0.0 (0.0) | 0.0 |
| Left_foot-R_5kHz | | 280.6 ± 36.6 (279.8) | 261.3 ± 35.6 (260.6) | 1.1e-81 |
| Left_foot-X_5kHz | | -9.6 ± 3.7 (-9.4) | -8.6 ± 2.2 (-8.4) | 2.2e-47 |
| Left_foot-R_50kHz | | 253.9 ± 33.1 (253.0) | 238.0 ± 32.8 (236.8) | 1.9e-66 |
| Left_foot-X_50kHz | | -21.2 ± 4.6 (-21.2) | -19.1 ± 4.1 (-19.1) | 5.4e-71 |
| Left_foot-R_250kHz | | 233.0 ± 30.9 (231.6) | 218.2 ± 30.1 (217.0) | 3.3e-68 |
| Left_foot-X_250kHz | | -18.5 ± 3.4 (-18.4) | -17.2 ± 3.3 (-17.1) | 3.3e-44 |
| Left_foot-R_500kHz | | 226.8 ± 30.2 | 0.0 ± 0.0 (0.0) | 0.0 |
| | | (225.3) | | |
| Left_foot-X_500kHz | | -19.7 ± 4.1 (-19.4) | 0.0 ± 0.0 (0.0) | 0.0 |
| Right_arm-R_5kHz | | 375.0 ± 57.9 (372.8) | 361.5 ± 51.8 (360.5) | 2.5e-21 |
| Right_arm-X_5kHz | | -15.7 ± 10.0 (-15.3) | -13.8 ± 4.3 (-13.6) | 1.3e-39 |
| Right_arm-R_50kHz | | 332.8 ± 54.7 (331.2) | 323.9 ± 49.6 (323.3) | 6.6e-11 |
| Right_arm-X_50kHz | | -35.5 ± 6.2 (-35.0) | -33.8 ± 4.5 (-33.5) | 7.0e-39 |
| Right_arm-R_250kHz | | 294.9 ± 49.1 (293.9) | 290.2 ± 45.7 (289.6) | 0.00014 |
| Right_arm-X_250kHz | | -37.4 ± 8.6 (-36.6) | -41.1 ± 7.5 (-40.3) | 1.6e-64 |
| Right_arm-R_500kHz | | 287.3 ± 48.5 (286.3) | 0.0 ± 0.0 (0.0) | 0.0 |
| Right_arm-X_500kHz | | -43.1 ± 13.0 (-41.3) | 0.0 ± 0.0 (0.0) | 0.0 |
| Left_arm-R_5kHz | | 380.6 ± 59.3 (377.9) | 366.1 ± 51.9 (364.4) | 5.2e-24 |
| Left_arm-X_5kHz | | -15.4 ± 8.0 (-15.0) | -13.9 ± 4.9 (-13.6) | 3.1e-24 |
| Left_arm-R_50kHz | | 339.3 ± 56.0 (337.2) | 329.1 ± 49.8 (327.8) | 8.8e-14 |
| Left_arm-X_50kHz | | -35.4 ± 6.0 (-34.8) | -33.7 ± 4.5 (-33.5) | 1.5e-37 |
| Left_arm-R_250kHz | | 300.9 ± 50.3 (299.1) | 295.4 ± 46.1 (293.1) | 1.4e-05 |
| Left_arm-X_250kHz | | -38.5 ± 8.9 (-37.6) | -42.3 ± 7.9 (-41.3) | 1.3e-62 |
| Left_arm-R_500kHz | | 291.8 ± 49.4 (290.0) | 0.0 ± 0.0 (0.0) | 0.0 |
| Left_arm-X_500kHz | | -44.6 ± 13.6 (-42.6) | 0.0 ± 0.0 (0.0) | 0.0 |
| Both_legs-R_5kHz | | 559.3 ± 72.3 (557.2) | 523.0 ± 70.3 (522.0) | 2.8e-74 |
| Both_legs-X_5kHz | | -20.0 ± 5.0 (- | -17.3 ± 4.7 (- | 1.9e-86 |
| | | 19.6) | 16.8) | |
| Both_legs-R_50kHz | | 502.9 ± 65.5 (500.5) | 475.8 ± 64.3 (473.4) | 1.9e-51 |
| Both_legs-X_50kHz | | -45.1 ± 9.0 (-44.9) | -38.8 ± 7.9 (-38.8) | 8.7e-153 |
| Both_legs-R_250kHz | | 456.1 ± 59.7 (454.0) | 435.0 ± 58.9 (432.4) | 1.8e-38 |
| Both_legs-X_250kHz | | -39.2 ± 6.6 (-39.1) | -34.1 ± 5.6 (-34.1) | 2.0e-196 |
| Both_legs-R_500kHz | | 442.7 ± 58.2 (440.5) | 0.0 ± 0.0 (0.0) | 0.0 |
| Both_legs-X_500kHz | | -39.8 ± 9.5 (-39.5) | 0.0 ± 0.0 (0.0) | 0.0 |
| Dressing_Weight | | 1.0 ± 0.0 (1.0) | 1.0 ± 0.5 (1.0) | 0.026 |
| Body_fat_percentage | | 25.9 ± 8.2 (25.5) | 24.9 ± 8.4 (24.3) | 1.3e-05 |
| Fatty_constitution | | 15.4 ± 6.4 (14.6) | 14.9 ± 6.6 (14.1) | 0.0044 |
| Lean_body_mass | | 43.5 ± 9.0 (40.4) | 44.0 ± 8.9 (41.3) | 0.034 |
| Muscle_mass | | 41.1 ± 8.6 (38.0) | 41.6 ± 8.5 (38.8) | 0.032 |
| Muscle_score | | 0.1 ± 1.3 (0.0) | 0.1 ± 1.3 (0.0) | 0.32 |
| Estimated_Bone_Mass | | 2.4 ± 0.4 (2.4) | 2.4 ± 0.4 (2.4) | 0.096 |
| Body_water_content | | 31.0 ± 5.9 (29.4) | 31.6 ± 6.1 (30.3) | 0.00015 |
| Standard_weight | | 56.3 ± 6.5 (55.6) | 56.5 ± 6.4 (56.1) | 0.41 |
| Body_mass_index | | 4.5 ± 15.2 (3.1) | 4.1 ± 15.5 (2.5) | 0.32 |
| Internal_Fat_Level | | 7.8 ± 4.2 (7.0) | 9.0 ± 4.4 (9.0) | 6.3e-24 |
| Foot_stool | | 92.2 ± 7.5 (92.0) | 90.0 ± 6.3 (90.0) | 3.7e-35 |
| Basal_Metabolism_Determination | | 10.3 ± 3.3 (10.0) | 11.0 ± 3.4 (11.0) | 2.7e-16 |
| Right_Leg-Body_Fat_Percentage | | 27.8 ± 7.4 (28.6) | 26.3 ± 7.9 (26.9) | 3.5e-11 |
| Right_leg-fat_content | | 2.9 ± 1.0 (2.8) | 2.7 ± 1.0 (2.6) | 4.8e-09 |
| Right_leg-lean_mass | | 7.6 ± 1.9 (7.0) | 7.7 ± 1.9 (7.2) | 0.24 |
| Right_leg-muscle_mass | | 7.2 ± 1.8 (6.7) | 7.2 ± 1.8 (6.8) | 0.24 |
| Right_Leg-Body_Fat_Percentage_Score | | 0.1 ± 1.4 (0.0) | -0.1 ± 1.5 (0.0) | 5.1e-07 |
| Right_leg-muscle_mass_score | | -0.5 ± 1.4 (0.0) | -0.6 ± 1.4 (-1.0) | 0.091 |
| Grip_strength_Right_1st | | 31.7 ± 10.3 (29.0) | 28.8 ± 9.1 (26.9) | 4.0e-30 |
| Grip_strength_right_second | | 32.4 ± 10.3 (30.0) | 29.2 ± 9.1 (27.4) | 1.7e-34 |
| Grip_strength_left_1st | | 30.9 ± 10.1 (28.0) | 27.9 ± 9.0 (25.9) | 3.6e-33 |
| Grip_strength_left_second | | 30.8 ± 10.0 (28.0) | 27.9 ± 8.9 (25.9) | 6.4e-32 |
| LDL_cholesterol | | 116.3 ± 28.9 (115.0) | 121.7 ± 32.5 (120.0) | 5.6e-10 |
| Calcium (Ca) | | 9.5 ± 0.4 (9.5) | 4.7 ± 0.2 (4.7) | 0.0 |
| Health_Status_Medications_Hypertension_Medication | | 3476 (26.8%) | 575 (37.9%) | 1.2e-19 |
| Health_Status_Medications_Hyperlipidemia_Medication | | 1495 (11.5%) | 388 (25.8%) | 3.0e-54 |
| Health_Status_Medications_Diabetes_Medications | | 599 (4.6%) | 134 (8.8%) | 2.1e-12 |
| %Predicted_Value_V25 | | 65.2 ± 31.5 (58.9) | 72.7 ± 56.8 (60.0) | 9.4e-07 |
| FVC | | 3.4 ± 0.9 (3.2) | 3.1 ± 0.8 (3.0) | 1.3e-29 |
| Predicted_FVC | | 2.9 ± 0.6 (2.8) | 3.0 ± 0.7 (2.9) | 0.0039 |
| %Predicted_FVC | | 115.4 ± 17.8 (115.9) | 105.5 ± 17.4 (105.2) | 1.7e-85 |
| FEV1 | | 2.7 ± 0.8 (2.6) | 2.4 ± 0.6 (2.3) | 1.1e-69 |
| Predicted_FEV1 | | 2.4 ± 0.7 (2.3) | 2.4 ± 0.6 (2.3) | 0.65 |
| %Predicted_FEV1 | | 114.5 ± 20.4 (113.8) | 100.4 ± 18.1 (100.5) | 1.6e-143 |
| FEV1%G | | 80.6 ± 7.6 (81.0) | 77.0 ± 8.0 (77.0) | 1.9e-55 |
| PEF | | 6.7 ± 2.2 (6.3) | 5.9 ± 2.2 (5.7) | 3.0e-32 |
| %Predicted_V50 | | 83.9 ± 28.3 (82.7) | 90.5 ± 32.7 (88.0) | 3.6e-13 |
| BDHQ_Roots_&_Vegetables | | 30.4 ± 25.0 (24.6) | 34.2 ± 27.0 (27.1) | 2.3e-07 |
| BDHQ_Tomato | | 22.9 ± 25.3 | 41.1 ± 36.8 | 1.3e-71 |
| | | (11.4) | (28.5) | |
| BDHQ_Mushroom | | 10.4 ± 9.5 (9.2) | 9.2 ± 8.9 (5.2) | 1.2e-06 |
| BDHQ_Seaweed | | 11.8 ± 11.5 (9.9) | 11.1 ± 11.1 (6.3) | 0.022 |
| BDHQ_Pastry | | 21.1 ± 24.1 (11.5) | 18.3 ± 22.6 (10.0) | 9.5e-06 |
| BDHQ_Wagashi | | 8.1 ± 10.8 (3.8) | 9.0 ± 10.9 (7.1) | 0.0045 |
| BDHQ_Senbei | | 17.9 ± 17.8 (17.9) | 14.2 ± 15.2 (8.2) | 1.0e-17 |
| BDHQ_Ice_Cream | | 18.5 ± 26.5 (8.0) | 31.6 ± 36.9 (17.1) | 7.0e-39 |
| BDHQ_Citrus | | 11.6 ± 19.7 (6.0) | 11.7 ± 23.2 (0.0) | 0.82 |
| BDHQ_Kaki-Strawberry | | 8.6 ± 15.3 (6.0) | 8.8 ± 18.6 (0.0) | 0.74 |
| BDHQ_Fruits_Others | | 32.5 ± 36.8 (14.8) | 43.7 ± 38.8 (32.1) | 1.8e-25 |
| BDHQ_Mayonnaise | | 6.2 ± 5.2 (5.2) | 7.6 ± 6.3 (5.8) | 6.8e-16 |
| BDHQ_Bread | | 32.7 ± 27.4 (25.0) | 51.4 ± 30.0 (57.7) | 2.7e-103 |
| BDHQ_Soba | | 21.8 ± 26.4 (11.9) | 16.0 ± 21.6 (9.3) | 7.0e-21 |
| BDHQ_Udon | | 21.9 ± 24.3 (16.0) | 32.1 ± 30.7 (20.8) | 6.0e-34 |
| BDHQ_Ramen | | 23.7 ± 28.1 (16.0) | 13.4 ± 20.6 (8.4) | 6.3e-63 |
| BDHQ_Pasta | | 13.3 ± 15.0 (9.7) | 11.8 ± 14.0 (9.3) | 6.9e-05 |
| BDHQ_Green_tea | | 145.1 ± 177.5 (61.9) | 210.9 ± 222.7 (123.7) | 2.1e-27 |
| BDHQ_tea-Oolong_tea | | 43.7 ± 100.5 (10.0) | 44.3 ± 111.8 (0.0) | 0.83 |
| BDHQ_Coffee | | 225.6 ± 180.4 (150.0) | 274.7 ± 191.7 (375.0) | 1.6e-20 |
| BDHQ_Cola | | 73.1 ± 126.3 (15.4) | 94.3 ± 155.0 (28.6) | 3.6e-07 |
| BDHQ_100%juice | | 43.2 ± 76.9 (13.3) | 50.4 ± 90.2 (13.3) | 0.0031 |
| BDHQ_Sugar | | 2.6 ± 4.4 (0.0) | 2.1 ± 4.4 (0.0) | 4.6e-05 |
| BDHQ_Meshi | | 326.7 ± 157.3 (312.0) | 283.2 ± 155.7 (260.0) | 9.1e-24 |
| BDHQ_Miso_soup | | 169.4 ± 121.6 (124.7) | 103.0 ± 109.3 (69.3) | 3.9e-95 |
| BDHQ_Sake | | 13.6 ± 52.0 (0.0) | 12.4 ± 48.6 (0.0) | 0.39 |
| BDHQ_Beer | | 117.2 ± 248.3 (0.0) | 91.0 ± 205.6 (0.0) | 7.0e-06 |
| BDHQ_Shochu | | 18.1 ± 47.2 (0.0) | 14.4 ± 38.4 (0.0) | 0.00056 |
| BDHQ_Whisky | | 2.9 ± 16.2 (0.0) | 1.3 ± 9.4 (0.0) | 1.5e-08 |
| BDHQ_Wine | | 4.9 ± 24.7 (0.0) | 3.6 ± 22.9 (0.0) | 0.044 |
| BDHQ_Raw_fish | | 23.0 ± 24.5 (15.6) | 26.7 ± 26.7 (20.4) | 3.7e-07 |
| BDHQ_Grilled_fish | | 45.7 ± 39.1 (36.2) | 37.8 ± 32.6 (29.0) | 2.7e-17 |
| BDHQ_Boiled_fish | | 53.0 ± 51.7 (33.4) | 47.8 ± 45.6 (33.4) | 4.3e-05 |
| BDHQ_Tempura&Fried_Fish | | 18.5 ± 19.7 (13.6) | 25.9 ± 22.6 (21.3) | 3.7e-33 |
| BDHQ_Yakiniku | | 14.5 ± 18.7 (9.7) | 16.1 ± 16.8 (11.9) | 0.00083 |
| BDHQ_Hamburg | | 28.7 ± 25.4 (21.3) | 32.9 ± 26.7 (26.8) | 1.1e-08 |
| BDHQ_Fried_food | | 24.3 ± 21.3 (18.2) | 29.9 ± 24.4 (23.5) | 6.2e-17 |
| BDHQ_Fry | | 60.7 ± 38.9 (56.0) | 55.4 ± 36.3 (52.3) | 1.4e-07 |
| BDHQ_Boiled_food | | 96.3 ± 72.3 (83.1) | 92.6 ± 66.6 (82.9) | 0.047 |
| BDHQ_Men_Soup | | 95.2 ± 78.9 (72.2) | 86.6 ± 70.9 (70.0) | 1.5e-05 |
| BDHQ_Soy_Sauce_quantity | | 1.6 ± 0.5 (1.6) | 1.5 ± 0.4 (1.5) | 1.9e-20 |
| BDHQ_Citrus_Season | | 11.4 ± 12.2 (8.0) | 19.6 ± 15.8 (18.6) | 3.7e-77 |
| BDHQ_Oshi_Season | | 6.8 ± 10.8 (1.7) | 14.9 ± 13.4 (9.3) | 3.0e-99 |
| BDHQ_Strawberry_Season | | 8.3 ± 11.9 (2.7) | 9.1 ± 12.5 (5.0) | 0.018 |
| BDHQ_Cooked_Salt | | 3.4 ± 1.2 (3.3) | 3.3 ± 1.3 (3.2) | 0.10 |
| BDHQ_Cooking_oil | | 10.9 ± 5.4 (10.2) | 12.0 ± 5.7 (11.5) | 1.0e-12 |
| BDHQ_Cooking_sugar | | 3.1 ± 1.9 (2.9) | 2.9 ± 1.8 (2.6) | 0.00017 |
| BDHQ_Low_fat_milk | | 33.8 ± 63.8 (0.0) | 40.5 ± 70.7 (0.0) | 0.00052 |
| BDHQ_Normal_breast | | 69.7 ± 80.7 (48.2) | 88.9 ± 92.5 (58.9) | 2.8e-14 |
| BDHQ_Chicken | | 24.0 ± 21.6 (15.7) | 26.4 ± 24.2 (24.6) | 0.00018 |
| BDHQ_Pork&Beef | | 32.7 ± 23.3 (30.8) | 36.0 ± 26.0 (32.0) | 3.5e-06 |
| BDHQ_Ham | | 8.4 ± 8.1 (4.9) | 9.1 ± 9.4 (5.1) | 0.0054 |
| BDHQ_lever | | 1.0 ± 3.0 (0.0) | 1.3 ± 2.8 (0.0) | 0.0083 |
| BDHQ_Squid-Octopus-Shrimp-Shellfish | | 15.4 ± 16.5 (11.8) | 13.3 ± 13.3 (7.8) | 3.1e-08 |
| BDHQ_Fish_with_bones | | 8.1 ± 14.0 (4.6) | 14.1 ± 19.6 (7.9) | 3.2e-30 |
| BDHQ_Tuna_can | | 3.5 ± 5.4 (2.8) | 4.0 ± 5.7 (3.1) | 0.0070 |
| BDHQ_Dried_fish | | 20.1 ± 21.2 (12.6) | 21.7 ± 21.1 (13.2) | 0.0040 |
| BDHQ_Oily_fish | | 19.0 ± 19.9 (13.1) | 18.8 ± 19.0 (13.1) | 0.72 |
| BDHQ_Fish_with_less_fat | | 20.3 ± 19.4 (13.7) | 16.8 ± 16.4 (13.1) | 4.6e-14 |
| BDHQ_egg | | 37.9 ± 25.2 (28.3) | 42.1 ± 28.4 (32.7) | 3.5e-08 |
| BDHQ_Tofu_deep-fried_tofu | | 46.7 ± 37.0 (38.4) | 52.5 ± 38.8 (40.3) | 3.9e-08 |
| BDHQ_Natto | | 21.9 ± 18.5 (16.7) | 13.2 ± 17.3 (6.4) | 3.5e-69 |
| BDHQ_potato | | 34.9 ± 32.2 (23.1) | 37.1 ± 35.3 (23.1) | 0.019 |
| BDHQ_Pickles_Green_leafy_vegetables | | 8.0 ± 10.2 (3.8) | 10.0 ± 11.3 (7.6) | 3.9e-10 |
| BDHQ_Pickles_Others | | 9.1 ± 11.8 (3.8) | 11.4 ± 12.7 (8.2) | 2.8e-11 |
| BDHQ_Raw_Lettuce_and_Cabbage | | 25.1 ± 21.2 | 29.1 ± 23.2 | 4.8e-10 |
| | | (19.0) | (20.9) | |
| BDHQ_Green_leafy_vegetables | | 32.6 ± 33.6 (25.9) | 29.7 ± 31.3 (23.0) | 0.00072 |
| BDHQ_Cabbage | | 32.1 ± 28.2 (28.8) | 37.2 ± 31.9 (29.9) | 7.3e-09 |
| BDHQ_Carrot&Pumpkin | | 17.1 ± 15.6 (14.8) | 19.8 ± 16.2 (17.1) | 1.7e-09 |
| BDHQ_Radish&turnip | | 15.5 ± 18.1 (9.9) | 15.1 ± 17.5 (9.9) | 0.34 |
| BDHQ_Zinc | | 8.2 ± 2.8 (7.8) | 8.4 ± 2.9 (8.0) | 0.068 |
| BDHQ_Copper | | 1.2 ± 0.4 (1.1) | 1.1 ± 0.4 (1.1) | 5.7e-07 |
| BDHQ_Manganese | | 2.9 ± 1.1 (2.8) | 3.0 ± 1.2 (2.9) | 0.00051 |
| BDHQ_Retinol | | 386.7 ± 433.9 (272.1) | 463.3 ± 413.4 (340.9) | 2.6e-11 |
| BDHQ_β-carotene_equivalent | | 2989.8 ± 2178.7 (2477.2) | 3277.2 ± 2207.3 (2825.0) | 2.2e-06 |
| BDHQ_retinol_equivalent | | 638.9 ± 504.7 (539.9) | 739.7 ± 486.3 (644.3) | 9.1e-14 |
| BDHQ_Vitamin_D | | 14.8 ± 11.3 (11.9) | 16.8 ± 12.5 (13.2) | 6.4e-09 |
| BDHQ_alpha_tocopherol | | 6.9 ± 2.8 (6.5) | 7.7 ± 3.0 (7.4) | 4.9e-24 |
| BDHQ_Vitamin_K | | 323.3 ± 176.5 (292.9) | 282.0 ± 173.4 (233.7) | 9.0e-18 |
| BDHQ_Vitamin_B1 | | 0.7 ± 0.3 (0.7) | 0.8 ± 0.3 (0.7) | 1.4e-13 |
| BDHQ_Vitamin_B2 | | 1.2 ± 0.5 (1.2) | 1.3 ± 0.5 (1.3) | 3.5e-12 |
| BDHQ_Niacin | | 17.3 ± 7.3 (16.1) | 18.3 ± 7.7 (17.1) | 4.9e-07 |
| BDHQ_Vitamin_B6 | | 1.2 ± 0.5 (1.1) | 1.3 ± 0.5 (1.2) | 0.0013 |
| BDHQ_Vitamin_B12 | | 10.2 ± 7.0 (8.5) | 10.9 ± 7.1 (9.1) | 0.0013 |
| BDHQ_Folic_Acid | | 298.0 ± 136.8 (275.8) | 317.7 ± 141.0 (297.4) | 3.3e-07 |
| BDHQ_Pantothenic_acid | | 6.4 ± 2.3 (6.1) | 6.6 ± 2.4 (6.3) | 0.010 |
| BDHQ_Vitamin_C | | 86.1 ± 49.9 (75.7) | 99.1 ± 55.0 (89.8) | 7.1e-18 |
| BDHQ_saturated_fatty_acids | | 13.8 ± 5.7 (13.1) | 16.1 ± 6.4 (15.3) | 8.8e-38 |
| BDHQ_monounsaturated_fatty_acid | | 18.5 ± 7.1 (17.6) | 20.9 ± 7.9 | 2.2e-27 |
| | | | (20.1) | |
| BDHQ_Polyunsaturated_fatty_acids | | 13.2 ± 4.8 (12.7) | 14.1 ± 5.2 (13.5) | 7.8e-10 |
| BDHQ_Cholesterol | | 368.6 ± 176.0 (343.8) | 414.3 ± 193.8 (394.3) | 7.8e-18 |
| BDHQ_Soluble_dietary_fiber | | 2.8 ± 1.3 (2.6) | 2.8 ± 1.3 (2.6) | 0.88 |
| BDHQ_insoluble_dietary_fiber | | 8.2 ± 3.3 (7.7) | 8.1 ± 3.4 (7.6) | 0.67 |
| BDHQ_Total_dietary_fiber | | 11.4 ± 4.7 (10.6) | 11.3 ± 4.8 (10.5) | 0.63 |
| BDHQ_Salt_Equivalent | | 11.1 ± 3.8 (10.5) | 11.0 ± 3.9 (10.5) | 0.63 |
| BDHQ_Sucrose | | 12.2 ± 9.1 (10.1) | 12.9 ± 9.0 (10.8) | 0.0024 |
| BDHQ_Daizein | | 16.1 ± 10.0 (14.3) | 13.2 ± 9.8 (10.8) | 1.4e-27 |
| BDHQ_Genistein | | 27.3 ± 16.9 (24.1) | 22.4 ± 16.5 (18.3) | 2.2e-26 |
| BDHQ_n-3_fatty_acids | | 2.8 ± 1.3 (2.5) | 2.9 ± 1.4 (2.7) | 3.7e-06 |
| BDHQ_n-6_fatty_acids | | 10.4 ± 3.7 (10.0) | 11.2 ± 4.0 (10.7) | 2.0e-10 |
| BDHQ_C04S | | 163.7 ± 122.0 (138.3) | 210.1 ± 139.9 (196.6) | 2.2e-33 |
| BDHQ_C06S | | 103.0 ± 78.6 (84.6) | 136.1 ± 91.4 (129.0) | 2.4e-39 |
| BDHQ_C08S | | 110.4 ± 92.3 (85.0) | 160.2 ± 119.9 (135.8) | 2.6e-51 |
| BDHQ_C10S | | 175.4 ± 130.4 (148.7) | 239.1 ± 159.5 (210.8) | 2.1e-47 |
| BDHQ_C10M | | 13.9 ± 10.4 (11.7) | 18.4 ± 12.1 (17.4) | 1.2e-42 |
| BDHQ_C12S | | 448.3 ± 395.0 (329.3) | 661.9 ± 525.0 (494.6) | 1.3e-49 |
| BDHQ_C14S | | 1085.1 ± 581.1 (982.3) | 1336.4 ± 674.9 (1215.7) | 2.4e-41 |
| BDHQ_C14M | | 78.2 ± 41.4 (71.9) | 94.3 ± 47.5 (88.4) | 1.7e-34 |
| BDHQ_C15S | | 99.8 ± 51.3 | 117.9 ± 56.4 | 3.9e-31 |
| | | (91.7) | (108.9) | |
| BDHQ_C15M | | 0.0 ± 0.0 (0.0) | 0.0 ± 0.0 (0.0) | nan |
| BDHQ_C16S | | 8224.1 ± 3199.8 (7808.7) | 9353.2 ± 3540.8 (8914.9) | 7.0e-31 |
| BDHQ_C16M | | 790.7 ± 378.1 (726.5) | 875.4 ± 421.1 (808.9) | 1.7e-13 |
| BDHQ_C163n6 | | 11.9 ± 9.2 (9.4) | 12.2 ± 9.2 (9.9) | 0.35 |
| BDHQ_C17S | | 131.1 ± 58.6 (122.0) | 147.6 ± 64.7 (138.6) | 1.2e-20 |
| BDHQ_C17M | | 91.2 ± 42.5 (84.5) | 103.4 ± 47.8 (95.9) | 1.0e-20 |
| BDHQ_C18S | | 2961.6 ± 1189.8 (2803.9) | 3371.6 ± 1319.2 (3220.7) | 1.8e-29 |
| BDHQ_C18M | | 16572.0 ± 6332.5 (15834.7) | 18785.4 ± 7032.3 (18091.6) | 3.6e-30 |
| BDHQ_C182n6 | | 10155.2 ± 3601.4 (9749.6) | 10818.1 ± 3931.3 (10406.6) | 6.6e-10 |
| BDHQ_C18n3 | | 1602.9 ± 604.2 (1540.5) | 1704.5 ± 673.7 (1645.9) | 3.0e-08 |
| BDHQ_C183n6 | | 7.8 ± 5.6 (6.3) | 9.8 ± 6.7 (8.5) | 3.5e-29 |
| BDHQ_C184n3 | | 91.7 ± 76.2 (68.5) | 94.4 ± 75.0 (71.9) | 0.20 |
| BDHQ_C20S | | 151.2 ± 56.1 (144.6) | 169.8 ± 62.0 (165.1) | 1.1e-27 |
| BDHQ_C20M | | 526.3 ± 325.7 (436.4) | 545.5 ± 326.0 (468.8) | 0.032 |
| BDHQ_C202n6 | | 45.1 ± 21.1 (41.9) | 50.4 ± 23.8 (47.3) | 4.8e-16 |
| BDHQ_C203n6 | | 29.4 ± 13.2 (27.6) | 33.4 ± 14.5 (31.3) | 2.2e-23 |
| BDHQ_C204n3 | | 34.9 ± 28.2 (26.3) | 35.7 ± 27.9 (27.5) | 0.27 |
| BDHQ_C204n6 | | 169.2 ± 78.4 (158.4) | 186.6 ± 83.6 (177.6) | 2.9e-14 |
| BDHQ_C205n3 | | 347.4 ± 282.7 (265.2) | 372.6 ± 290.9 (285.9) | 0.0015 |
| BDHQ_C22S | | 79.1 ± 29.5 | 86.4 ± 32.4 | 3.2e-16 |
| | | (76.0) | (84.0) | |
| BDHQ_C22M | | 373.1 ± 335.8 (272.6) | 371.0 ± 323.8 (280.7) | 0.82 |
| BDHQ_C222n6 | | 0.0 ± 0.0 (0.0) | 0.0 ± 0.0 (0.0) | nan |
| BDHQ_C225n3 | | 102.1 ± 76.2 (80.2) | 106.1 ± 76.8 (85.2) | 0.056 |
| BDHQ_C225n6 | | 9.3 ± 7.3 (7.1) | 9.3 ± 7.2 (7.3) | 0.91 |
| BDHQ_C226n3 | | 582.9 ± 433.5 (460.9) | 622.3 ± 442.9 (499.5) | 0.0012 |
| BDHQ_C24S | | 33.4 ± 12.7 (32.0) | 36.6 ± 13.9 (35.5) | 8.2e-18 |
| BDHQ_C24M | | 54.0 ± 38.6 (43.3) | 56.3 ± 38.8 (45.4) | 0.033 |
| BDHQ_alpha-carotene | | 327.2 ± 293.8 (283.5) | 379.2 ± 304.2 (325.7) | 5.1e-10 |
| BDHQ_$\beta$-carotene | | 2726.3 ± 2018.4 (2239.7) | 2975.9 ± 2035.8 (2542.0) | 8.3e-06 |
| BDHQ_Cryptoxanthin | | 194.3 ± 213.2 (116.7) | 211.6 ± 247.8 (115.8) | 0.0098 |
| BDHQ_$\beta$-tocopherol | | 0.4 ± 0.1 (0.4) | 0.4 ± 0.1 (0.4) | 4.2e-07 |
| BDHQ_gamma-tocopherol | | 13.0 ± 4.8 (12.5) | 14.0 ± 5.3 (13.5) | 7.2e-12 |
| BDHQ_delta-tocopherol | | 3.4 ± 1.3 (3.2) | 3.3 ± 1.3 (3.2) | 0.57 |
| BDHQ_C07S | | 0.8 ± 0.8 (0.6) | 1.0 ± 0.9 (0.7) | 5.7e-12 |
| BDHQ_C13S | | 2.3 ± 2.4 (1.6) | 2.9 ± 2.8 (2.0) | 9.5e-13 |
| BDHQ_C15SA | | 23.5 ± 17.8 (19.5) | 30.3 ± 20.4 (28.3) | 7.9e-34 |
| BDHQ_C16SI | | 11.7 ± 8.9 (9.6) | 15.1 ± 10.3 (14.1) | 6.9e-34 |
| BDHQ_C17SA | | 23.9 ± 17.8 (20.4) | 31.0 ± 20.6 (29.2) | 1.2e-35 |
| BDHQ_C162 | | 12.5 ± 10.2 (9.4) | 13.3 ± 10.6 (10.4) | 0.0026 |
| BDHQ_C164 | | 10.7 ± 9.6 (7.5) | 11.5 ± 10.0 (8.3) | 0.0027 |
| BDHQ_C215N3 | | 10.2 ± 9.2 (7.2) | 10.7 ± 9.3 (7.8) | 0.038 |
| BDHQ_C224N6 | | 6.8 ± 3.8 (6.2) | 7.2 ± 3.9 (6.8) | 2.5e-05 |
| Estimated_energy_requirements | | 2146.9 ± 302.2 | 2093.1 ± 271.1 | 1.6e-12 |
| | | (1987.5) | (1982.5) | |
| BDHQ_energy | | 1883.9 ± 593.9 (1801.1) | 1909.4 ± 610.0 (1817.8) | 0.13 |
| BDHQ_weight | | 2150.6 ± 708.7 (2058.3) | 2244.7 ± 747.3 (2170.5) | 4.1e-06 |
| BDHQ_water | | 1742.1 ± 604.5 (1664.3) | 1834.0 ± 639.2 (1761.0) | 1.5e-07 |
| BDHQ_Protein | | 70.5 ± 26.7 (66.1) | 73.4 ± 28.4 (68.8) | 0.00018 |
| BDHQ_Animal_protein | | 39.9 ± 20.1 (36.3) | 44.2 ± 21.8 (40.2) | 3.6e-13 |
| BDHQ_Plant_protein | | 30.6 ± 10.2 (29.3) | 29.2 ± 10.1 (27.9) | 2.3e-07 |
| BDHQ_Lipid | | 52.5 ± 19.6 (50.0) | 58.7 ± 21.5 (56.1) | 8.3e-26 |
| BDHQ_Animal_Lipids | | 24.6 ± 11.5 (22.8) | 28.7 ± 12.9 (26.9) | 5.4e-31 |
| BDHQ_Plant_lipids | | 27.9 ± 10.5 (26.7) | 30.0 ± 11.3 (29.1) | 1.2e-11 |
| BDHQ_Carbohydrates | | 254.3 ± 85.7 (242.7) | 249.3 ± 88.0 (237.4) | 0.037 |
| BDHQ_Ash | | 18.3 ± 6.1 (17.4) | 18.8 ± 6.6 (18.0) | 0.0016 |
| BDHQ_Sodium | | 4375.9 ± 1486.7 (4172.7) | 4355.4 ± 1555.9 (4164.4) | 0.63 |
| BDHQ_potassium | | 2350.9 ± 925.7 (2207.8) | 2531.9 ± 993.5 (2396.6) | 2.8e-11 |
| BDHQ_Calcium | | 505.8 ± 235.6 (467.1) | 583.6 ± 259.9 (551.6) | 1.9e-27 |
| BDHQ_Magnesium | | 251.9 ± 92.4 (239.2) | 259.1 ± 96.3 (247.0) | 0.0061 |
| BDHQ_Lynn | | 1048.2 ± 400.8 (984.0) | 1124.3 ± 433.6 (1056.6) | 1.4e-10 |
| BDHQ_Iron | | 7.6 ± 3.0 (7.1) | 7.7 ± 3.1 (7.2) | 0.23 |
| HbA1c_NGSP | | 5.7 ± 0.6 (5.6) | 5.8 ± 0.5 (5.7) | 1.9e-14 |
| eGFR | | 79.9 ± 15.7 (79.1) | 66.3 ± 12.6 (66.0) | 5.5e-252 |
| Onset_in_1yr_DM | | Onset: 258 (2.5%) | Onset: 12 (3.8%) | 0.19 |
| | | No Onset: 10101 | No Onset: 300 | |
| Onset_in_1yr_DysL | | Onset: 1508 (32.8%) No Onset: 3093 | Onset: 22 (23.7%) No Onset: 71 | 0.081 |
| Onset_in_1yr_HTN | | Onset: 1139 (17.1%) No Onset: 5510 | Onset: 33 (18.4%) No Onset: 146 | 0.72 |

## Supplementary Note 6. Disease Diagnostic Criteria

**Supplementary Table 2. Disease Diagnostic Criteria**

| Disease | Diagnostic Criteria |
|---|---|
| Diabetes | Having an HbA1c value of 6.5% or higher, a fasting blood sugar level of 126 mg/dL or above, or being under treatment with anti-diabetic medication[1]. |
| Dyslipidemia | Defined by Japanese guidelines[2] as having an LDL cholesterol level of 120 mg/dL or above, an HDL cholesterol level below 40 mg/dL, a triglyceride level of 150 mg/dL or above, or currently receiving medication treatment for the condition. |
| Hypertension | Having a systolic blood pressure of 140 mmHg or higher, diastolic blood pressure of 90 mmHg or higher, or being on anti-hypertensive medication. Although the latest ACC/AHA guidelines[3] have lowered the diagnostic threshold to 130/80 mmHg, this study retained the traditional Japanese guideline[4] of 140/90 mmHg due to the inclusion of data prior to 2017 and for consistency with Japan's standards. |

## Supplementary Note 7. Items for Disease Onset Prediction Model

**Supplementary Table 3. Items Each Disease Onset Prediction Model Used**

| Diabetes | Dyslipidemia | Hypertension |
|---|---|---|
| HbA1c_NGSP | HbA1c_NGSP | Systolic_blood_pressure |
| Serum_glucose | Total_cholesterol | Diatolic_blood_pressure |
| TG_triglycerides | TG_triglycerides | Serum_glucose |
| MCH | HDL_cholesterol | Age |
| Sleep_Drowsiness_At_Work | LDL_cholesterol | Height |
| RbaPWV | RbaPWV | RbaPWV |
| LbaPWV | LbaPWV | LbaPWV |
| eGFR | Urea_nitrogen | eGFR |
| MCV | BMI | Right_Leg_Body_Fat_Percentage |
| Calcium | Body_mass_score | Right_Arm-Body_Fat_Percentage_Score |
| Red_blood_cell_count | Torso-lean_mass | BDHQ_Soy_Sauce_quantity |
| BDHQ_Fried_food | AST_GOT | BDHQ_Daizein |
| BDHQ_Pickles_Green_leafy_vegetables | Right_leg-muscle_mass | BDHQ_Vitamin_B2 |
| BDHQ_Radish&turnip | BDHQ_Plant_lipids | BDHQ_Miso_soup |
| BDHQ_Chicken | Right_arm-R_500kHz | BDHQ_retinol_equivalent |
| BDHQ_17M | Right_arm-X_5kHz | BDHQ_Boiled_fish |
| BDHQ_16M | Left_foot-X_5kHz | Left_half-R_250kHz |
| BDHQ_Cryptoxanthin | Left_arm-R_500kHz | Both_legs-R_5kHz |
| BDHQ_Pantothenic_acid | Left_arm-X_500kHz | %Predicted_FVC |
| Predicted_FEV1 | %Predicted_FVC | Predicted_FVC |

Items beginning with “BDHQ”[5–7] refer to estimated dietary intake values, and those ending in "Hz" are impedance values measured for various body parts using a body composition analyzer.

## Supplementary Note 8. XGBoost Hyperparameter Search

**Supplementary Table 4. XGBoost Parameter Grid Search Options**

| n_estimators | max_depth | min_child_weight |
|---|---|---|
| 50 | 2 | 1 |
| 100 | 4 | 2 |
| 200 | 6 | 3 |

## Supplementary Note 9. Propensity score matching with age

Dataset shift refers to the phenomenon where there is a discrepancy between the distributions of the training data and the test data. In Fig. 5 of the main text, we present the results of the ODROP method in cases where it is unknown which specific item distributions are shifted. However, in real-world scenarios, it is conceivable that domain knowledge and other information might allow the identification of shifted items. Therefore, in this section, we examine the ODROP method in situations where it is known that specific items have undergone a shift.

We selected age as the known item experiencing a shift because the distribution of age included significant shift (Fig. 2 of the text). Subsequently, we organized the datasets by performing propensity score matching using logistic regression with age as the covariate. Finally, we applied the ODROP method to the matched Wakayama dataset. The evaluation was conducted using the rejection rate, as detailed in the "External Validation using Wakayama Health Checkups" and "Methods" section of the main text.

The statistics adjusted through propensity score matching is presented in Supplementary Table 5 below.

**Supplementary Table 5. Propensity Score Matched Age Statistics (Mean±std)**

| Target Diseases | Age: Hirosaki data | Age: Wakayama data |
|---|---|---|
| Diabetes | 56.8 ± 11.5 | 55.1 ± 14.1 |
| Dyslipidemia | 54.0 ± 12.4 | 51.6 ± 15.4 |
| Hypertension | 53.1 ± 10.8 | 50.3 ± 13.5 |

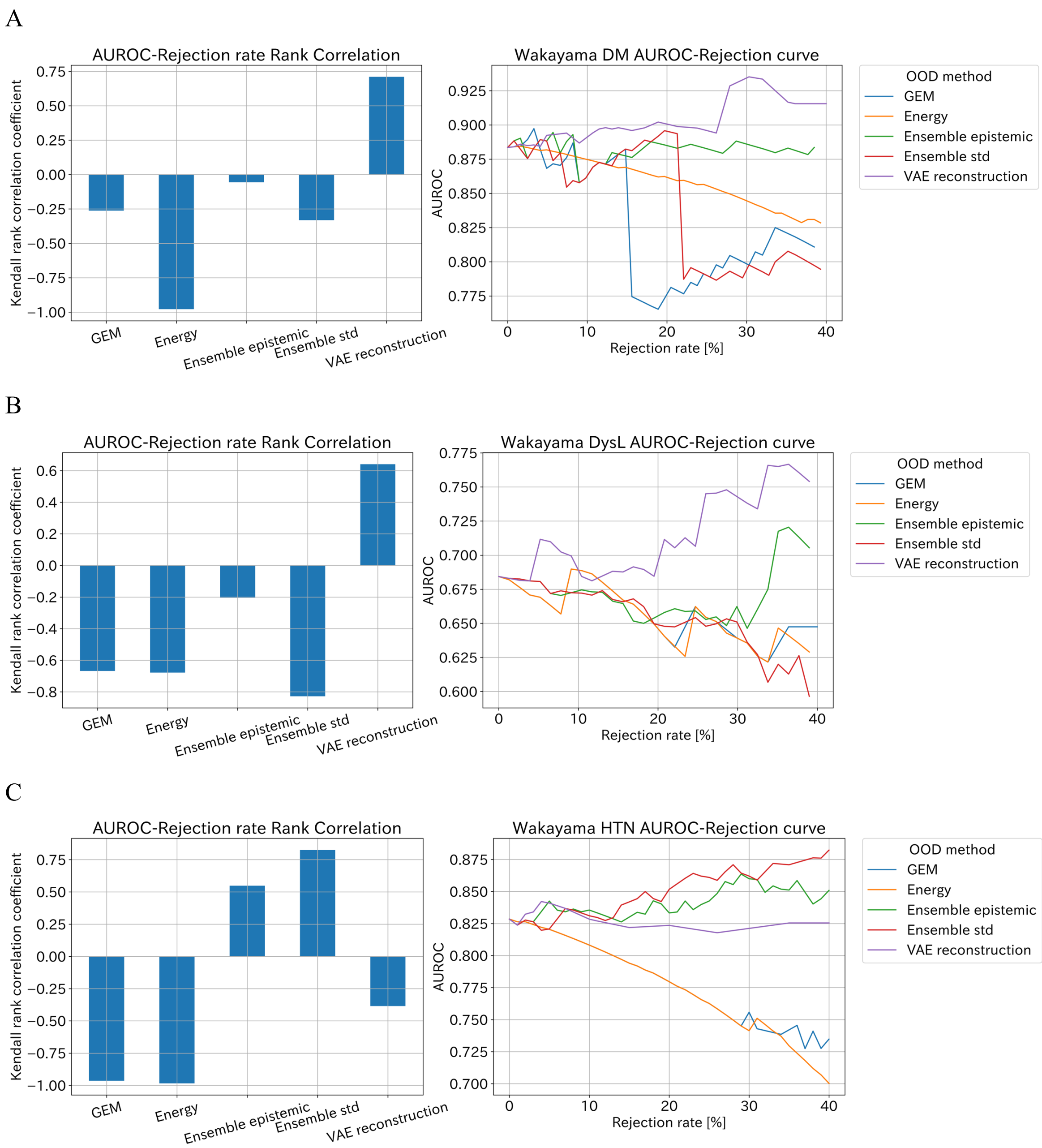

**Supplementary Fig. 5 AUROC-rejection rate rank correlation coefficients and AUROC-rejection curves in Wakayama Health Checkup data**

**A: Diabetes Melius (DM), B: Dyslipidemia (DysL), C: Hypertension (HTN).**

The AUROC for diabetes and hypertension at a rejection rate of 0% post-matching were 0.084 and 0.828, respectively, reflecting increases of 0.094 and 0.017 from pre-matching Wakayama dataset. This improvement can be attributed to the matching, which helped correct for age data distribution shift.

VAE reconstruction was the only OOD detection method with a positive rank correlation coefficient in Supplementary Fig. 5 A and B, showing a stable improvement in AUROC through the rejection curve. In C, ensemble epistemic and ensemble std had positive coefficients, with the rejection curve confirming an upward trend in AUROC. These results are same to that of without propensity score matching (Fig. 5 of the main text). Therefore, it was confirmed that the ODROP method remains effective even when shifts in a specific item such as age are known and adjusted through matching. This indicates that the proposed method captures shifts in factors beyond age, including clinical practices and environmental factors, contributing to improved predictive performance.

## Supplementary Reference